\documentclass[lettersize,journal]{IEEEtran}
\ifCLASSOPTIONcompsoc
  \usepackage[nocompress]{cite}
\else
  \usepackage{cite}
\fi

\usepackage{algorithm}
\usepackage{algpseudocode}
\usepackage{amsmath}
\usepackage{amssymb}
\usepackage{amsfonts}
\usepackage{graphicx}
\usepackage{subfigure}
\usepackage{xcolor}
\usepackage{hyperref}
\ifCLASSINFOpdf
\else
\fi

\begin{document}
\title{Sketched Multi-view Subspace Learning for \\ Hyperspectral Anomalous Change Detection}

\author{Shizhen~Chang,~\IEEEmembership{Member,~IEEE,}
        Michael~Kopp,
        Pedram~Ghamisi,~\IEEEmembership{Senior Member,~IEEE}
\thanks{Manuscript received **; revised **.} 
\IEEEcompsocitemizethanks{\IEEEcompsocthanksitem S. Chang and M. Kopp are with the Institute of Advanced Research in Artificial Intelligence, Landstraßer Hauptstraße 5, 1030 Vienna, Austria (e-mail: shizhen.chang@iarai.ac.at; michael.kopp@iarai.ac.at).
\IEEEcompsocthanksitem P. Ghamisi is the Institute of Advanced Research in Artificial Intelligence, Landstraßer Hauptstraße 5, 1030 Vienna, Austria, and also with the Helmholtz-Zentrum Dresden-Rossendorf (HZDR), Helmholtz Institute Freiberg for Resource Technology (HIF), Machine Learning Group, Chemnitzer Str. 40, D-09599 Freiberg, Germany (e-mail: pedram.ghamisi@iarai.ac.at).}} 

\markboth{IEEE Transactions on GEOSCIENCE AND REMOTE SENSING}%
{Shell \MakeLowercase{\textit{et al.}}: A Sample Article Using IEEEtran.cls for IEEE Journals}

\IEEEtitleabstractindextext{%
\begin{abstract}
In recent years, multi-view subspace learning has been garnering increasing attention. It aims to capture the inner relationships of the data that are collected from multiple sources by learning a unified representation. In this way, comprehensive information from multiple views is shared and preserved for the generalization processes. As a special branch of temporal series hyperspectral image (HSI) processing, the anomalous change detection task focuses on detecting very small changes among different temporal images. However, when the volume of datasets is very large or the classes are relatively comprehensive, existing methods may fail to find those changes between the scenes, and end up with terrible detection results. In this paper, inspired by the sketched representation and multi-view subspace learning, a sketched multi-view subspace learning (SMSL) model is proposed for HSI anomalous change detection. The proposed model preserves major information from the image pairs and improves computational complexity by using a sketched representation matrix. Furthermore, the differences between scenes are extracted by utilizing the specific regularizer of the self-representation matrices. To evaluate the detection effectiveness of the proposed SMSL model, experiments are conducted on a benchmark hyperspectral remote sensing dataset and a natural hyperspectral dataset, and compared with other state-of-the art approaches. The codes of the proposed method will be made available online\footnote{https://github.com/ShizhenChang/SMSL}.
\end{abstract}

}

\maketitle

\IEEEdisplaynontitleabstractindextext

%
\IEEEpeerreviewmaketitle

\ifCLASSOPTIONcompsoc
\IEEEraisesectionheading{\section{Introduction}\label{sec:introduction}}
\else
\section{Introduction}
\label{sec:introduction}
\fi

\IEEEPARstart{A}{s} ONE of the typical research fields of image processing, change detection focuses on measuring the progression of one or more patterns that perform substantially differently among multi-temporal images of the same scene \cite{acito2017introductory}. In response to the demands of diverse disciplines, change detection has been widely applied for remote sensing \cite{Francesca2007Theoretical, wang2018getnet, du2019unsupervised}, video surveillance \cite{collins2000introduction, stauffer2000learning}, medical diagnosis and treatment \cite{lemieux1998detection, grattarola2019change}, civil infrastructure \cite{taneja2015geometric, nagy2001volume}, and underwater sensing \cite{wang2021target, vivekananda2021multi}. And with the development of hyperspectral imaging technology \cite{ghamisi2017advances, chang2020subspace}, utilizing the wealth of spectral information together with spatial correlations between the objects to capture changes in the multi-temporal HSIs has attracted increasing attention \cite{liu2019review}.

Generally speaking, the change detection task needs to squeeze the input image pairs into one output image that can reflect the potentially changed object with higher values. Our challenge is to distinguish real changes from the interference caused by sensor noise, camera motion, illumination variation, shadows, or atmospheric absorption \cite{radke2005image}. And based on the interests of the particular application, change detection can be categorized into two main topics, general change detection \cite{hou2021hyperspectral, eismann2008hyperspectral} and anomalous change detection (ACD) \cite{zhou2016novel, wu2018hyperspectral}. In this paper, we focus on exploring the potential of the anomalous changes in HSIs. What we are interested in are those relatively small and unusual changes within the broad changed or unchanged backgrounds, such as small objects that suddenly appear, disappear, or move. 

\begin{figure}
    \centering
    \includegraphics[width=0.5\textwidth]{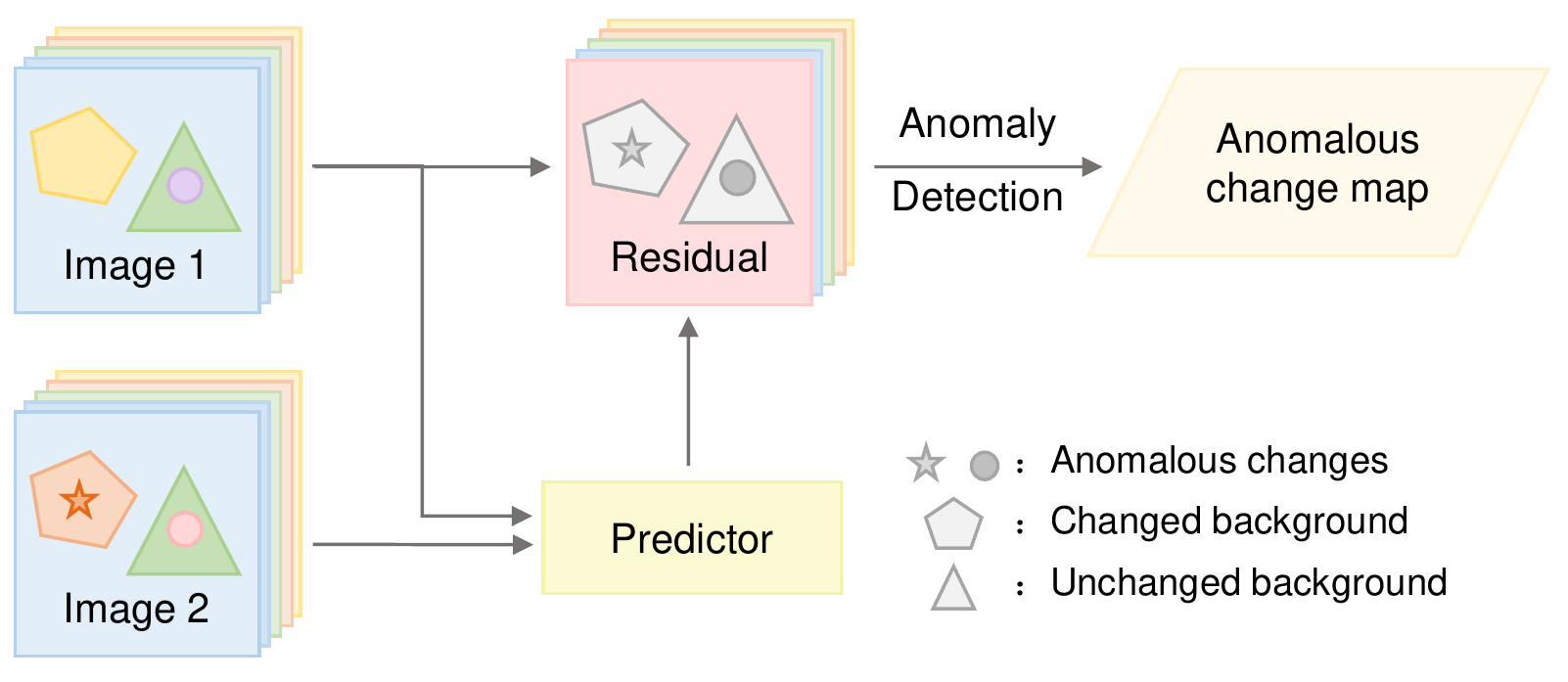}
    \caption{The basic diagram of anomalous change detection.}
    \label{fig:1}
\end{figure}
\begin{figure*}[t]
    \centering
    \includegraphics[width=0.93\textwidth]{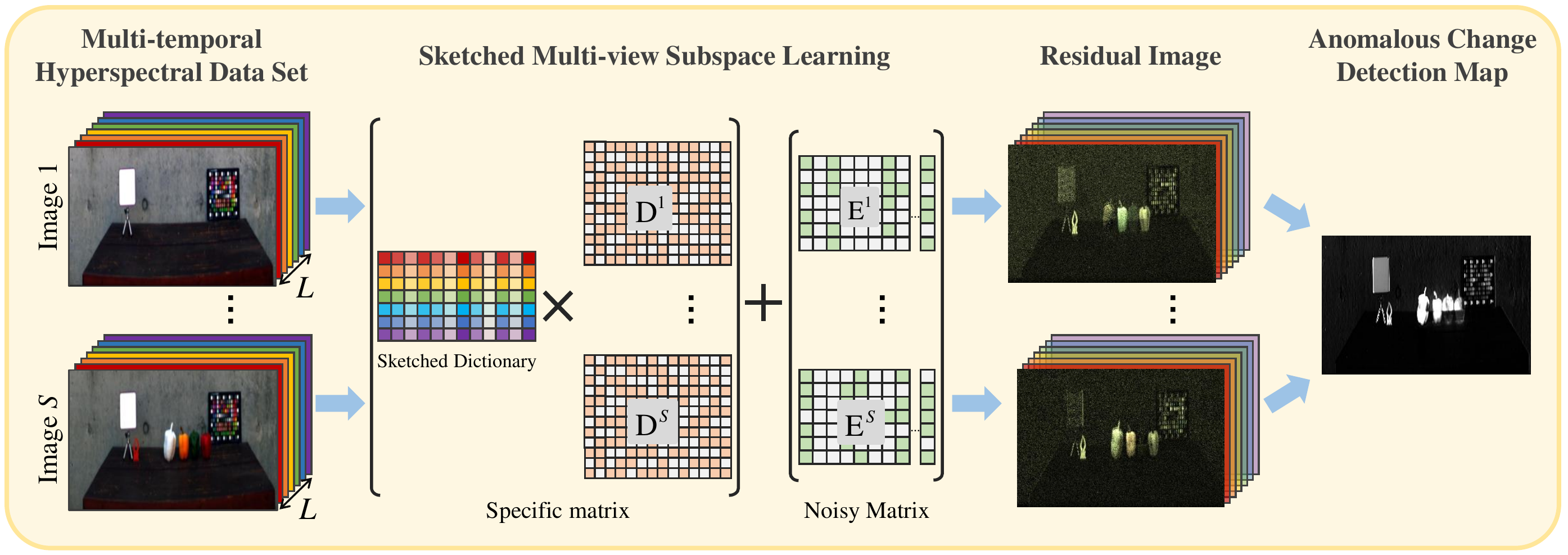}
    \caption{The flowchart of the proposed SMSL model.}
    \label{fig2}
\end{figure*}
In order to detect those small objects, the basic idea of ACD algorithms is to define a predictor to minimize the spectral difference of the backgrounds while highlighting the changes in very high-dimensional multi-temporal images \cite{eismann2007hyperspectral}. A simplified diagram of ACD is shown in Fig. 1. In recent years, a number of ACD algorithms have been proposed in the literature to cope with this problem. One effective predictor is the chronochrome (CC) \cite{schaum2015theoretical, meola2008image} method, which uses the joint second-order statistic between two images to capture small changes. The spectral differences in backgrounds are modeled by the least square linear regression. Another widely used method, covariance equalization (CE) \cite{meola2008image, schaum2004hyperspectral}, assumes the images have the same statistical distribution after the whitening/dewhitening transform. Based on nonlinear Gaussian distribution, the cluster kernel Reed-Xiaoli (CKRX) algorithm \cite{zhou2016novel} was proposed and applied for change detection. The CKRX method groups background pixels into clusters and then applies a fast eigendecomposition algorithm to generate the anomaly index. Focusing on Gaussian and elliptically contoured (EC) distribution, the kernel anomalous change detection algorithm was proposed \cite{padron2019kernel}, which extends the detection process to nonlinear counterparts based on the theory of reproducing kernels’ Hilbert space. Traditional models assume the distribution of land covers obeys a statistical model, and then deduce the detection output through statistical assumptions.

With the development of signal processing and machine learning theories, more detectors have been proposed. Wu et. al. proposed a slow feature analysis-based method to explore potential small changes by assuming the background signals are invariant slow varying features \cite{wu2015hyperspectral}. The top bands of the residual images that are highly related to changes are finally selected as the input for RX anomaly detection \cite{reed1990adaptive}. Based on the sparse representation theory, a joint sparse representation anomalous change detection method \cite{wu2018hyperspectral} was proposed. For every test pixel, the surrounding pixels in its dual window are sparsely represented by a randomly selected stacked background dictionary matrix, and the detection output is determined by active bases of the dictionary. To better capture the nonlinear features from bitemporal images, an autoencoder-based two-Siamese network was proposed \cite{hu2021hyperspectral}, which utilizes bidirectional predictors to minimize the reconstruction errors between the image pairs. To fully explore the feature correlations of the images, a self-supervised hyperspectral spatial-spectral feature understanding network (HyperNet) was proposed \cite{hu2022hypernet}, which achieves pixel-level features for change detection. Unlike traditional predictors, the machine/deep-learning-based models focus on capturing the feature differences between backgrounds and anomalous changes, to extract the changed objects from multi-temporal images. However, those slight differences among the temporal images are still very difficult to be perfectly represented.

Nowadays, multi-view data analysis has gained increasing attention in many real-world applications, since data are usually collected from diverse domains or obtained from various feature subsets \cite{zhang2020deep, dhillon2011multi}. To better combine the consensus and complementary information among multiple views, the model is designed to give a comprehensive understanding and improve generalization performance \cite{guo2021rank}. Among all the topics related to multi-view learning, subspace learning is one of the most typical, which aims to obtain a latent subspace to align features for the inputs \cite{ zhao2017multi, chen2021adaptive}. Representative methods are successfully utilized for classification \cite{you2019multi} and clustering \cite{yin2018multiview, zhang2020tensorized}. However, for large-volume data, the huge complexity and memory consumption of the multi-view learning methods will cause a serious computational burden. On the other hand, traditional multi-view learning methods pay more attention to homogeneous information of each view and, thus, will ignore the correlation among the views.

For the proposed hyperspectral anomalous change detection task, instead of capturing the most common information among the views, we are interested in extracting small abnormal objects and excluding them from the background instances. To address this issue, a sketched multi-view subspace learning (SMSL) model with a consistent constraint and a specific constraint is proposed in this paper. The flowchart of the proposed SMSL model is shown in Fig. \ref{fig2}, which illustrates its three main steps. First, considering the large volume of the hyperspectral images, a sketched dictionary is calculated from the union matrix of all the images. Then the residual fractions between the neighboring views corresponding to the specific matrix and the noisy matrix are obtained. And finally, the anomalous change detection map is derived by calculating the norm of the residuals. Our main contributions are summarized as follows:

\begin{enumerate}
\item[--] It is the first attempt to combine multi-view subspace learning with change detection to distinguish small anomalous changes from temporal images. Considering the large volume of hyperspectral datasets and high computational consumption of the optimization process, a sketched dictionary is utilized to preserve most of the information from the original data with much smaller sample size.

\item[--] With a low-rank regularizer that constrains the consistent part of the coefficient matrix and two regularizers that constrain the specific part, the SMSL model guarantees the common information of the data are the lowest-rank represented, while the differences are maximally separated.

\item[--] Experiments conducted on sufficient HSI datasets demonstrate the effectiveness of the proposed method. Detailed analysis concludes the convergence of the model and the performance related to the parameters.
\end{enumerate}

\section{Related Works}
\subsection{Subspace Learning}
Subspace learning mainly focuses on recovering the subspace structure of the data. Recently, self-representative methods, which assume data instances can be approximately formed by a combination of other instances of the data, have been used for clustering and classification tasks. The representation models are generated via structured convex regularizers. For example, the sparse subspace clustering (SSC) \cite{elhamifar2013sparse} method calculates data clusters in a low-dimensional subspace using the $\ell_1$-norm. For a given data $\mathrm{\bf X}\in\mathbb{R}^{d\times N}$, where $d$ represents the dimension and $N$ represents the total number of samples of the data, the objective function of SSC is written as:
\begin{equation}
    \min||\mathrm{Z}||_1 \ \ \mathrm{s.t.} \ \mathrm{\bf X}=\mathrm{\bf X}\mathrm{Z}, \ \mathrm{diag}(\mathrm{Z})=0,
\end{equation}
where $\mathrm{Z}\in\mathbb{R}^{N\times N}$ is the coefficient matrix. 

Based on the low-rank representation model, the LRR \cite{liu2012robust} objective was proposed to solve the following problem:
\begin{equation}
    \min\limits_{\mathrm{Z},\mathrm{E}}||\mathrm{Z}||_\star+\lambda||\mathrm{E}||_{2,1} \ \ \mathrm{s.t.} \ \mathrm{\bf X}=\mathrm{\bf X}\mathrm{Z}+\mathrm{E},
\end{equation}
where the coefficient matrix $\mathrm{Z}$ is low-ranked in this case and $\mathrm{E}\in\mathbb{R}^{d\times N}$ is the error matrix corresponds to the sample-specific corruptions.

\subsection{Multi-view Subspace Learning}
Given that real-world data are usually collected from multiple sources or represent different feature types, multi-view subspace learning methods are proposed to learn a common subspace of different views. For instance, Guo \cite{guo2013convex} proposed a convex subspace learning model which jointly solves the optimization problem and learns the common subspace using a sparsity inducing norm. Ding et al. \cite{ding2016robust} proposed a robust multi-view subspace learning algorithm that uses dual low-rank decomposition and two supervised graph regularizers to obtain the view-invariant subspace.

Specifically, by jointly exploring consistency and specificity for subspace representation learning, Luo et al. \cite{luo2018consistent} design a novel multi-view self-representation model for clustering. 

Let $\mathrm{\bf X}^s\in\mathbb{R}^{d_s\times N}$ be the $s$-th view of all data, where $d_s$ denotes the dimension of $\mathrm{\bf X}$; the multi-view self-representation model can be formulated as:
\begin{equation}\label{eq1}
    \mathrm{\bf X}^s=\mathrm{\bf X}^s\mathrm{Z}^s+\mathrm{E}^s,
\end{equation}
where $\mathrm{Z}^s$ and $\mathrm{E}^s$ correspond to the coefficient matrix and the error matrix of $\mathrm{\bf X}^s$, respectively.

By arguing that the coefficient matrix of different views contains  a consistency term $\mathrm{C}$ and a view-specific term $\mathrm{D^s}$, then Eq. (\ref{eq1}) can be rewritten as:
\begin{equation}
    \mathrm{\bf X}^s=\mathrm{\bf X}^s(\mathrm{C}+\mathrm{D}^s)+\mathrm{E}^s.
\end{equation}

To better excavate the common information and ensure the unification property among all the views, the nuclear norm is imposed to constrain the consistent matrix and the $\ell_2$-norm is chosen for the specific matrix. Thus, the model is optimized via the Augmented Lagrange Multiplier (ALM) and the finally clustering result is derived by spectral clustering.

\section{Methodology}
The idea of multi-view subspace learning leads to new perspectives for multi-temporal data analysis. However, the optimization process of multi-view subspace learning models includes inverse derivations of the data itself, so the good performance is usually compromised by a very high price of computational complexity. And more importantly, directly applying the multi-view subspace learning model to detect anomalous changes cannot fully consider the abnormal information among the views. In this paper, to better extract specific fractions and effectively solve the subspace learning model, a sketched multi-view subspace learning model is proposed for hyperspectral anomalous change detection.

\subsection{Sketched Dictionary}
As has been introduced in Section II, existing subspace learning models use data itself as the dictionary to find the subspaces, but the large consumption of both storage and processing time makes it hard to handle large-scale datasets.

For the proposed hyperspectral anomalous change detection problem, which has abundant spectral information and relatively large data volume, the existing multi-view based subspace learning models are infeasible for deriving a detection result in a computationally efficient manner. Thus, to compress images into a scalable size and preserve main information, the sketched dictionary of the inputs is designed.

Assume the hyperspectral dataset $\{\mathrm{\bf X}^s\in\mathbb{R}^{L\times N}\}$ have $s$ phases in total, where $s\in\{1,2,...,S\}$, and $L$ and $N$ are the dimension and number of pixels of the $s$-th phase, respectively. In order to preserve most of the common information and enlarge differences of the coefficients among the views, we define a sketching matrix $R\in\mathbb{R}^{N\times N_\mathrm{H}}$ to obtained the sketched dictionary $\mathrm{H}\in\mathbb{R}^{L\times N_\mathrm{H}}$:
\begin{equation}
       \mathrm{H}=[\mathrm{\bf X}^1,\mathrm{\bf X}^2,\cdots,\mathrm{\bf X}^S]R.
\end{equation}

The sketching matrix is intended to compress data while retaining as much information as possible. It has been proved that the sketched representation can largely reduce the computational complexity, meanwhile, preserve the data information \cite{traganitis2017sketched, zhai2020nonlocal}. In particular, defining a random matrix using Johnson-Lindenstrauss transforms (JLT) can hold this property. In this paper, the random JLT matrix is independent and identically distributed (i.i.d.) with the values of the entries drawn from $\mathcal{N}(0,1)$ normal distribution scaled by the factor $1/\sqrt{N_\mathrm{H}}$.

\subsection{Sketched Multi-view Subspace Learning}
After obtaining the sketched dictionary $\mathrm{H}$, the sketched multi-view subspace learning model is proposed for anomalous change detection. For each phase of the data $\mathrm{\bf X}^s$, assuming it can be approximated formulated by the matrix multiplication of the sketched dictionary and corresponding coefficient matrix and a noisy matrix, the expressive function can be derived as follows:
\begin{equation}\label{eq2}
    \mathrm{\bf X}^s=\mathrm{H}\mathrm{Z}^s+\mathrm{E}^s,
\end{equation}
where $\mathrm{Z}^s$ and $\mathrm{E}^s$ denote the coefficient matrix and noisy matrix corresponding to $\mathrm{\bf X}^s$. Then, combining this with the consistency and specificity property of the coefficient matrix, Eq. (\ref{eq2}) can be modeled as:
\begin{equation}
    \mathrm{\bf X}^s=\mathrm{H}(\mathrm{C}+\mathrm{D}^s)+\mathrm{E}^s.
\end{equation}

Fig. \ref{fig:SMSL} displays the sketched multi-view subspace learning model with consistency and specificity where the common information shared through the consistent part uses the most representative information from the dictionary, and the differences of the specific parts among the views are large. Generally, the objective function of the SMSL model is expressed as:

\begin{figure}
    \centering
    \includegraphics[width=0.5\textwidth]{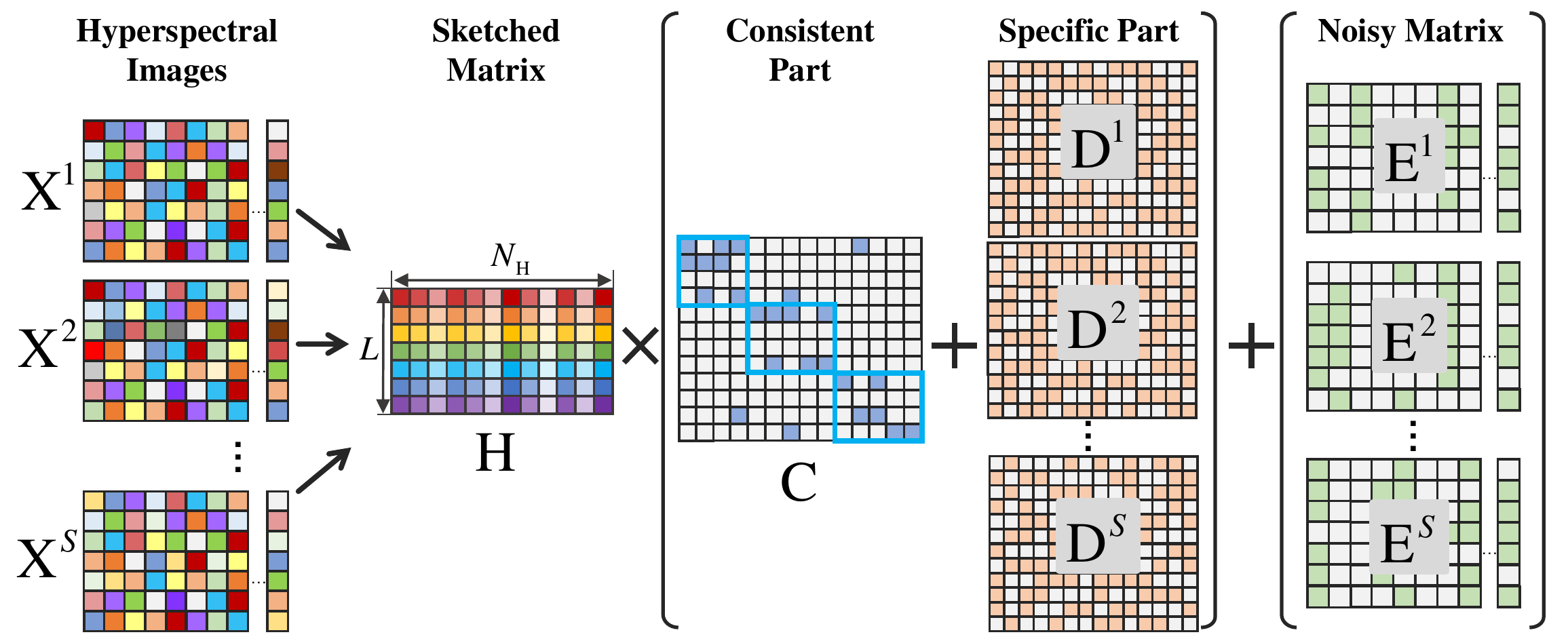}
    \caption{The basis model of SMSL. For each input image, its coefficient matrix is constructed by a consistent coefficient matrix $\textrm{C}$ and a specific coefficient matrix $\textrm{D}^s$, where the consistent matrix preserves the most common information among all the phases.}
    \label{fig:SMSL}
\end{figure}

\begin{equation*}
    \begin{split}
       \min_{\mathrm{E}^s,\mathrm{C},\mathrm{D}^s}  &\  \sum_s\mathcal{L}^s(\mathrm{\bf X}^s,\mathrm{H}(\mathrm{C}+\mathrm{D}^s))+\lambda_1\Omega_\mathrm{C}\\
        & +\lambda_2\sum\Omega_{\mathrm{D}^s} +\lambda_3\sum_s\mathcal{F}_{s,t\neq s}(\mathrm{D}^s,\mathrm{D}^t),
    \end{split}
\end{equation*}
where $\lambda_1, \lambda_2, \lambda_3>0$ are three trade-off parameters to balance the four terms, $\sum_s\mathcal{L}^s(\mathrm{\bf X}^s,\mathrm{H}(\mathrm{C}+\mathrm{D}^s))$ is the total loss of subspace representation, and the reconstructed coefficient matrices $\mathrm{C}$ and $\mathrm{D}^s$ are regularized with $\Omega_\mathrm{C}$ and $\Omega_{\mathrm{D}^s}$. $\mathcal{F}_s(\mathrm{D}^s,\mathrm{D}^t)$ measures the difference between $\mathrm{D}^s$ and $\mathrm{D}^t$. In our work, we want the difference between between $\mathrm{D}^s$ and $\mathrm{D}^t$ as large as possible, so the relaxed exclusivity is utilized to verify the similarity of the matrices.\\
{\bf Definition 1. (\textit{Relaxed Exclusivity \cite{guo2016exclusivity, wang2017exclusivity}})} \textit{The definition of relaxed exclusivity between $\mathbf{U}\in\mathbb{R}^{m\times n}$ and $\mathbf{V}\in\mathbb{R}^{m\times n}$ is $\mathcal{F}(\mathbf{U},\mathbf{V})=||\mathbf{U}\odot \mathbf{V}||_1=\sum_{i,j}|\mathbf{u}_{ij}\cdot \mathbf{v}_{ij}|$, where $||\cdot||_1$ is the $\ell_1$ norm, $|\cdot|$ represents the absolute value, and $\odot$ designates the Hadamard product which operates an element-wise multiplication of the two matrices.
}

It is noted that minimizing the relaxed exclusivity term can guarantee the two matrices are as orthogonal as possible. More specifically, the performance of detecting anomalies is strongly related to the comprehensive understanding of the common part and the identification of the differences. Thus, we design the proposed SMSL model with multiple regularization terms as:
\begin{equation}\label{eq8}
    \begin{split}
           \min_{\mathrm{E}^s,\mathrm{C},\mathrm{D}^s}  &\  \sum_{s=1}^S ||\mathrm{E}^s||_{2,1}+\lambda_1||\mathrm{C}||_\star+\frac{\lambda_2}{2}\sum_{s=1}^S||\mathrm{D}^s||_F^2 \\
        & +\lambda_3\sum_{s=1}^S\sum_{t\neq s}||\mathrm{D}^{t}\odot \mathrm{D}^s||_1\\
        \textrm{s.t.} \ \ \ & \ \mathrm{\bf X}^s=\mathrm{H}(\mathrm{C}+\mathrm{D}^s)+\mathrm{E}^s, (\mathrm{C}+\mathrm{D}^s)^\top\textbf{1}=\mathbf{1},
    \end{split}
\end{equation}
where $||\cdot||_{2,1}$ is a $\ell_{2,1}$-norm that ensures that the columns of the matrix are sparse\footnote{For a matrix $A\in\mathbb{R}^{m\times n}$, the definition of its $\ell_{2,1}$-norm is: $||A||_{2,1}=\sum_{j=1}^n\sqrt{\sum_{i=1}^m a_{i,j}^2}$.}, $||\cdot||_\star$ is the nuclear norm that ensures that the matrix is low-rank, and $||\cdot||_F$ is the Frobenius norm. By adding a sum-to-one constraint to the columns of the coefficient matrix, the images are assumed to be absolutely represented by the representative model and the consistent part is more robust to anomalies.
\subsection{Optimization}
According to the objective function of our SMSL model in Eq. (\ref{eq8}), we can simultaneously obtain the subspace representation of multi-views and optimize the consistent matrix. To pursue the optimal solutions of all variables, the proposed model is divided into several subproblems, and the Augmented Lagrange Multiplier (ALM) algorithm is utilized.

By introducing two auxiliary variables $\mathrm{W}^s$ and $\mathrm{J}$ to replace $\mathrm{E}^s$ and $\mathrm{C}$, respectively, our model can be equivalently rewritten as:
\begin{equation}\label{eq9}
    \begin{split}
        \min_{\mathrm{E}^s,\mathrm{W}^s,\mathrm{C},\mathrm{J},\mathrm{D}^s} & \  \sum_{s=1}^S ||\mathrm{W}^s||_{2,1}+\lambda_1||\mathrm{J}||_\star+\frac{\lambda_2}{2}\sum_{s=1}^S||\mathrm{D}^s||_F^2 \\
        & +\lambda_3\sum_{s=1}^S\sum_{t\neq s}||\mathrm{D}^{t}\odot \mathrm{D}^s||_1 \\
        \textrm{s.t.} \ \ \ \ \ \ \ & \ \mathrm{\bf X}^s=\mathrm{H}(\mathrm{C}+\mathrm{D}^s)+\mathrm{E}^s, \mathrm{E}^s=\mathrm{W}^s, \\
        & \ (\mathrm{C}+\mathrm{D}^s)^\top\textbf{1}=\mathbf{1}, \mathrm{C}=\mathrm{J}.
    \end{split}
\end{equation}
Then, the augmented Lagrange function is formulated as:
\begin{equation}
    \begin{split}
       & \mathcal{L}(\mathrm{C},\mathrm{J},\mathrm{D}^s,\mathrm{E}^s,\mathrm{W}^s,\mathrm{\bf Y}^s_1,\mathrm{\bf Y}^s_2,\mathrm{\bf Y}^s_3,\mathrm{\bf Y}_4)=\sum_{s=1}^S ||\mathrm{W}^s||_{2,1}\\
       &+\lambda_1||\mathrm{J}||_\star+\frac{\lambda_2}{2}\sum_{s=1}^S||\mathrm{D}^s||_F^2+\lambda_3\sum_{s=1}^S\sum_{t\neq s}||\mathrm{D}^{t}\odot \mathrm{D}^s||_1 \\ &+\frac{\mu}{2}\sum_{s=1}^S||\mathrm{\bf X}^s-\mathrm{H}(\mathrm{C}+\mathrm{D}^s)-\mathrm{E}^s+\frac{\mathrm{\bf Y}_1^s}{\mu}||_F^2 \\
       &+\frac{\mu}{2}\sum_{s=1}^S||\mathrm{\bf 1}^\top(\mathrm{C}+\mathrm{D}^s)-\mathrm{\bf 1}^\top+\frac{\mathrm{\bf Y}_2^s}{\mu}||_F^2 \\
       &+\frac{\mu}{2}\sum_{s=1}^S||\mathrm{E}^s-\mathrm{W}^s+\frac{\mathrm{\bf Y}_3^s}{\mu}||_F^2+\frac{\mu}{2}||\mathrm{C}-\mathrm{J}+\frac{\mathrm{\bf Y}_4}{\mu}||_F^2,
    \end{split}
\end{equation}
where $\{\mathrm{\bf Y}^s_1,\mathrm{\bf Y}^s_2,\mathrm{\bf Y}^s_3\}_{s\in[S]}$ and $\mathrm{\bf Y}_4$ are the Lagrange multipliers and $\mu>0$ is a penalty parameter. To optimize the above unconstrained function, we divide it into six subproblems and optimize each of them with other variables fixed, alternatively. The optimization process can be orgnized as follows:

{\bf 1) $\mathrm{C}$-subproblem}: Fixing the other variables, the variable $\mathrm{C}$ is optimized by the following subproblem:
\begin{equation*}
    \begin{split}
       & \mathrm{C}^\star=\mathop{\arg\min}\limits_{\mathrm{C}}\frac{\mu}{2}\sum_{s=1}^S||\mathrm{\bf X}^s-\mathrm{H}(\mathrm{C}+\mathrm{D}^s)-\mathrm{E}^s+\frac{\mathrm{\bf Y}_1^s}{\mu}||_F^2 \\
       &+\frac{\mu}{2}\sum_{s=1}^S||\mathrm{\bf 1}^\top(\mathrm{C}+\mathrm{D}^s)-\mathrm{\bf 1}^\top+\frac{\mathrm{\bf Y}_2^s}{\mu}||_F^2+\frac{\mu}{2}||\mathrm{C}-\mathrm{J}+\frac{\mathrm{\bf Y}_4}{\mu}||_F^2.
    \end{split}
\end{equation*}
By differentiating the objective function with respect to $\mathrm{C}$ and setting the derivation to zero, the update rule of $\mathrm{C}^\star$ is obtained:
\begin{equation*}
\begin{split}
  & \mathrm{C}^\star=A^{-1}B \\
   \mathrm{with} \ \ &A=S\mathrm{H}^\top\mathrm{H}+ S\mathrm{\bf 1}\mathrm{\bf 1}^\top+\textbf{I}, \\
  \mathrm{and} \ \ \  & B=\sum_{s=1}^S\mathrm{H}^\top(\mathrm{\bf X}^s-\mathrm{H}\mathrm{D}^s-\mathrm{E}^s+\frac{\mathrm{\bf Y}_1^s}{\mu}) \\
   &-\sum_{s=1}^S\mathrm{\bf 1}(\mathrm{\bf 1}^\top\mathrm{D}^s-\mathrm{\bf 1}^\top+\frac{\mathrm{\bf Y}_2^s}{\mu})+\mathrm{J}-\frac{\mathrm{\bf Y}_4}{\mu}.
\end{split}
\end{equation*}

{\bf 2) $\mathrm{J}$-subproblem}: With fixed variables, $\mathrm{J}$ can be optimized by the following problem: 
\begin{equation*}
       \mathrm{J}^\star=\mathop{\arg\min}\limits_{\mathrm{J}}\lambda_1||\mathrm{J}||_\star+\frac{\mu}{2}||\mathrm{C}-\mathrm{J}+\frac{\mathrm{\bf Y}_4}{\mu}||_F^2.
\end{equation*}
Note that the above function is equivalent to:
\begin{equation*}
       \mathrm{J}^\star=\mathop{\arg\min}\limits_{\mathrm{J}}\frac{\lambda_1}{\mu}||\mathrm{J}||_\star+\frac{1}{2}||\mathrm{J}-\mathrm{C}-\frac{\mathrm{\bf Y}_4}{\mu}||_F^2.
\end{equation*}
According to \cite{cai2010singular}, the approximation of the nuclear norm can be solved with the singular value thresholding (SVT) algorithm; the update rule is thus written as:
\begin{equation*}
       \mathrm{J}^\star=\mathrm{\bf U}\mathcal{D}_{\lambda_1/\mu}(\Sigma) \mathrm{\bf V}^\top,
\end{equation*}
where $\Sigma=\mathrm{C}+\frac{\mathrm{\bf Y}_4}{\mu}$, and the soft-thresholding operator $\mathcal{D}_\tau(\varepsilon)$ is defined as: 
\begin{equation*}
        \mathcal{D}_\tau(\varepsilon)=\max(\varepsilon-\tau,0)+\min(\varepsilon+\tau,0).
\end{equation*}

{\bf 3) $\mathrm{D}^s$-subproblem}: We can find that the variables $\{\mathrm{D}^s\}_{s\in[S]}$ are independent of each other, so for a fixed phase $s$, $\mathrm{D}^s$ can be solved by the following function with other variables fixed:
\begin{equation*}
\begin{split}
    \mathrm{D}^{s\star}&=\mathop{\arg\min}\limits_{\mathrm{D}^s}\frac{\lambda_2}{2}||\mathrm{D}^s||_F^2+\lambda_3\sum_{t\neq s}||\mathrm{D}^{t}\odot \mathrm{D}^s||_1 \\ &+\frac{\mu}{2}||\mathrm{\bf X}^s-\mathrm{H}(\mathrm{C}+\mathrm{D}^s)-\mathrm{E}^s+\frac{\mathrm{\bf Y}_1^s}{\mu}||_F^2 \\
    &+\frac{\mu}{2}||\mathrm{\bf 1}^\top(\mathrm{C}+\mathrm{D}^s)-\mathrm{\bf 1}^\top+\frac{\mathrm{\bf Y}_2^s}{\mu}||_F^2.
\end{split}
\end{equation*}

To calculate the optimal solution of $\mathrm{D}^s$, the key point is to obtain the partial derivation with respect to $\mathrm{D}^s$ for the $\ell_1$ norm of the Hardamart product function. \\
{\bf Lemma 1.} \textit{For finite dimensional matrices $\mathbf{U,V}\in \mathbb{R}^{m\times n}$, the partial derivative of the function $\mathcal{F}(\mathbf{U}, \mathbf{V})=||\mathbf{U}\odot\mathbf{V}||_1$ with respect to $\mathbf{V}$ is
\begin{equation*}
    \frac{\partial\mathcal{F}}{\partial\mathbf{V}}=|\mathbf{U}|\odot\mathrm{sign}(\mathbf{V}),
\end{equation*}
where $\mathrm{sign}(\cdot)$ is the component-wise sign function.\\
Proof.} The partial derivation with respect to each element $\mathbf{v}_{kl}$ of $\mathbf{V}$ can be written as:
\begin{equation*}
\begin{split}
    &\frac{\partial}{\partial \mathbf{v}_{kl}}(\sum_{i,j}|\mathbf{U}\odot\mathbf{V}|_{i,j}) =\frac{\partial}{\partial \mathbf{v}_{kl}}(\sum_{i,j}|\mathbf{u}_{ij}||\mathbf{v}_{ij}|) \\
    &=\sum_{i,j}|\mathbf{u}_{ij}|\frac{\partial|\mathbf{v}_{ij}|}{\partial \mathbf{v}_{kl}}=\sum_{i,j}|\mathbf{u}_{ij}|\delta_{ki}\delta_{lj}\mathrm{sign}(\mathbf{v}_{kl}) \\
    &=|\mathbf{u}_{kl}|\mathrm{sign}(\mathbf{v}_{kl}),
\end{split}
\end{equation*}
where the operator $\delta_{ki}=1$ when $k=i$, else, $\delta_{ki}=0$. So Lemma 1 holds.

For the current view $s$, assuming that the elements of $\mathrm{D}^{s}$ are all positive and following Lemma 1, the close form solution of $\mathrm{D}^s$ can be written as:
\begin{equation*}
\begin{split}
&\mathrm{D}^s=(\lambda_2\textbf{I}+\mu\mathrm{H}^\top\mathrm{H}+\mu\textbf{1}\textbf{1}^\top)^{-1}\times(-\lambda_3\sum_{t\neq s}|\mathrm{D}^t| \\ &+\mu\mathrm{H}^\top(\mathrm{\bf X}^s-\mathrm{H}\mathrm{C}-\mathrm{E}^s+\frac{\mathrm{\bf Y}_1^s}{\mu})-\mu\textbf{1}(\textbf{1}^\top\mathrm{C}-\textbf{1}^\top+\frac{\mathrm{\bf Y}_2^s}{\mu})),
\end{split}
\end{equation*}
where $\mathrm{D}^{t}$ can be conceptualized as constant matrices of the current view. And the update rule of $\mathrm{D}^{s\star}$ is
\begin{equation*}
\mathrm{D}^{s\star}_{ij}=\max\{\mathrm{D}^s_{ij},0\}.
\end{equation*}

{\bf 4) $\mathrm{E}^s$-subproblem}: By fixing other variables, the update rule of $\mathrm{E}^s$ is:
\begin{equation*}
\begin{split}
    \mathrm{E}^{s\star} & =\mathop{\arg\min}\limits_{\mathrm{E}^s}\frac{\mu}{2}||\mathrm{\bf X}^s-\mathrm{H}(\mathrm{C}+\mathrm{D}^s)-\mathrm{E}^s+\frac{\mathrm{\bf Y}_1^s}{\mu}||_F^2 \\
    &+\frac{\mu}{2}||\mathrm{E}^s-\mathrm{W}^s+\frac{\mathrm{\bf Y}_3^s}{\mu}||_F^2.
\end{split}\end{equation*}
Then the optimal value can be obtained by taking a partial derivation with respect to $\mathrm{E}^s$ and setting it to zero:
\begin{equation*}
  \mathrm{E}^{s\star}=\frac{1}{2}(\mathrm{\bf X}^s-\mathrm{H}(\mathrm{C}+\mathrm{D}^s)+\frac{\mathrm{\bf Y}_1^s}{\mu}+\mathrm{W}^s-\frac{\mathrm{\bf Y}_3^s}{\mu}).
\end{equation*}

\begin{algorithm}[t]
\caption{SMSL-ACD: Sketched Multi-view Subspace Learning for Hyperspectral Anomalous Change Detection}
\label{alg1}
\hspace*{0.02in} {\bf Input:}
The multi-temporal dataset $\mathrm{\bf X}$ $\{\mathrm{\bf X^s}\}_{s\in[S]}$, the sketched dictionary $\mathrm{\bf H}$, and the parameters  $\lambda_1$, $\lambda_2$, and $\lambda_3$.
\begin{algorithmic}[1]
\State Initialize coefficient matrices $\mathrm{C}=\mathrm{J}=\mathrm{D}^s=\mathrm{\mathbf{Y}}_4=0$, $\mathrm{E}^s=\mathrm{W}^s=\mathrm{\mathbf{Y}}_1^s=\mathrm{\mathbf{Y}}_3^s=0$, $\mathrm{\mathbf{Y}}_2^s=0$, set parameters $\mu=10^{-5}$, $\mu_{\max}=10^{5}$, $\rho=1.1$, maximum iteration times $iter=60$, and the stopping threshold $\epsilon=10^{-5}$.
\While{iterations$<iter$}
\State Update $\mathrm{C}$ and $\mathrm{J}$ according to subproblem 1)$-$2);
\For{$s\in[1,2,...,S]$}
\State Update $\mathrm{\textbf{D}}^s$, $\mathrm{\textbf{E}}^s$, and $\mathrm{\textbf{W}}^s$ according to subproblem 3)$-$5);
\State Update the multipliers $\mathrm{\mathbf{Y}}_1^s$, $\mathrm{\mathbf{Y}}_2^s$, and $\mathrm{\mathbf{Y}}_3^s$ according to subproblem 6);
\EndFor
\State Update the multipliers $\mathrm{\mathbf{Y}}_4$ and $\mu$ according to subproblem 6);
\State \textbf{If} converges \textbf{then}
\State \ \ \ \ break;
\EndWhile
\State \Return $\mathrm{\mathbf{D}}^s$ and $\mathrm{\mathbf{E}}^s$.
\end{algorithmic}
\hspace*{0.02in} {\bf Output:}
The specific matrices $\mathrm{H}\cdot\mathrm{D}^s$, noisy matrices $\mathrm{\mathbf{E}}^s$
\end{algorithm}

\begin{figure*}[h]
\centering
\subfigure[]{
\includegraphics[width=0.28\linewidth]{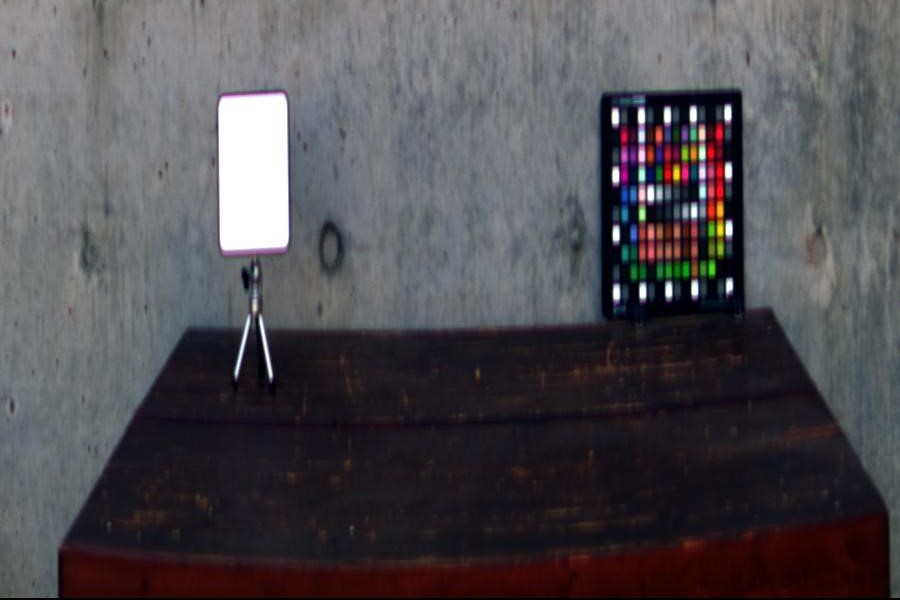}}
\subfigure[]{
\includegraphics[width=0.28\linewidth]{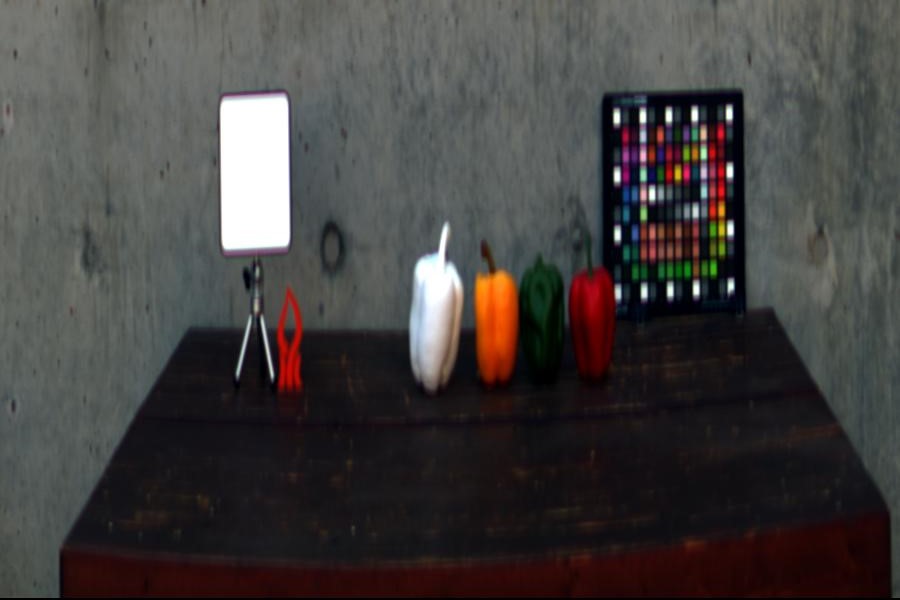}}
\subfigure[]{
\includegraphics[width=0.28\linewidth]{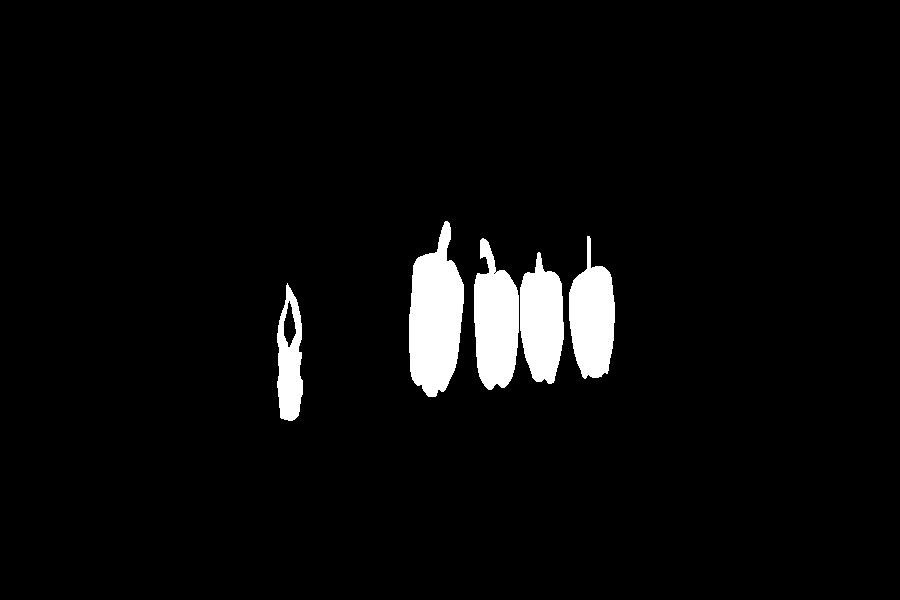}}
\caption{RGB-color images and the change mask of the Object$\_$1550$\_$1558 dataset: (a) Object$\_$1550, (b) Object$\_$1558, and (c) change mask.}
\label{object15501558}
\end{figure*}

\begin{figure*}[htp]
\centering
\subfigure[]{
\includegraphics[width=0.17\linewidth]{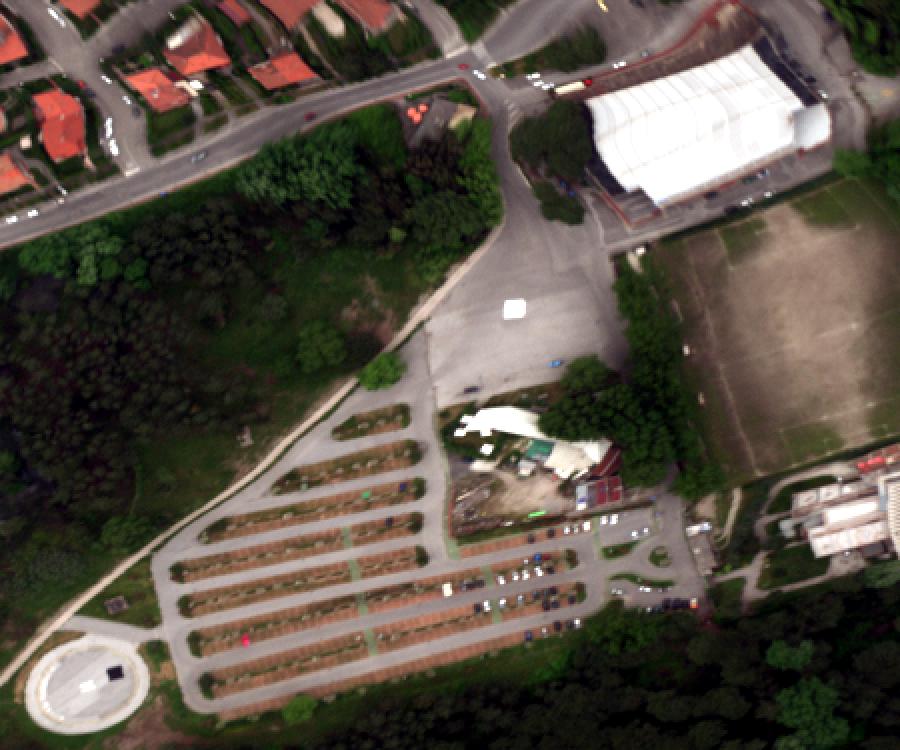}}
\subfigure[]{
\includegraphics[width=0.17\linewidth]{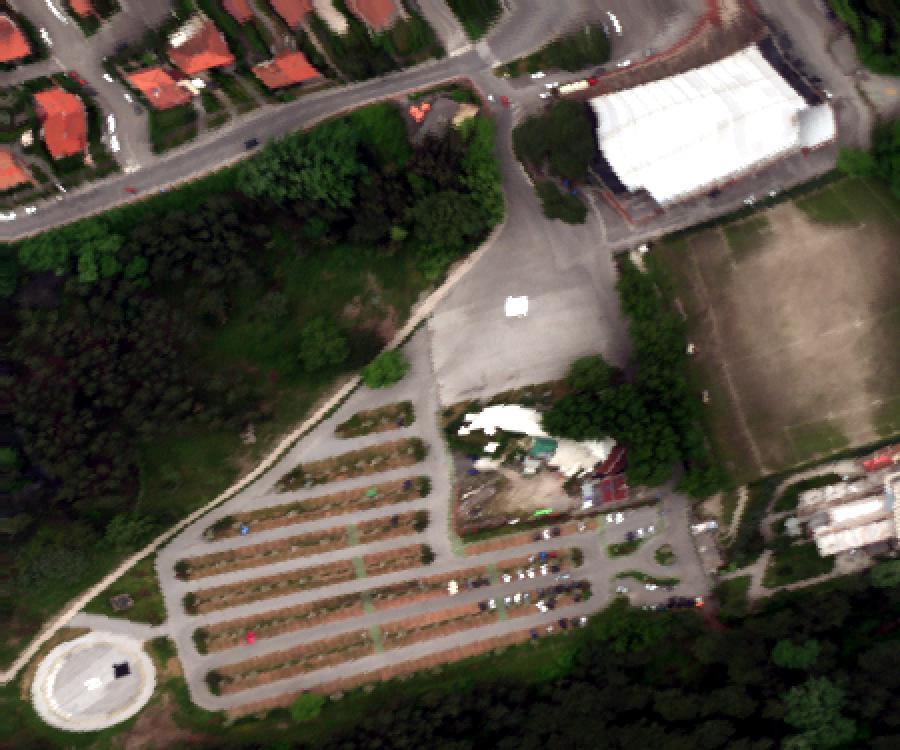}}
\subfigure[]{
\includegraphics[width=0.17\linewidth]{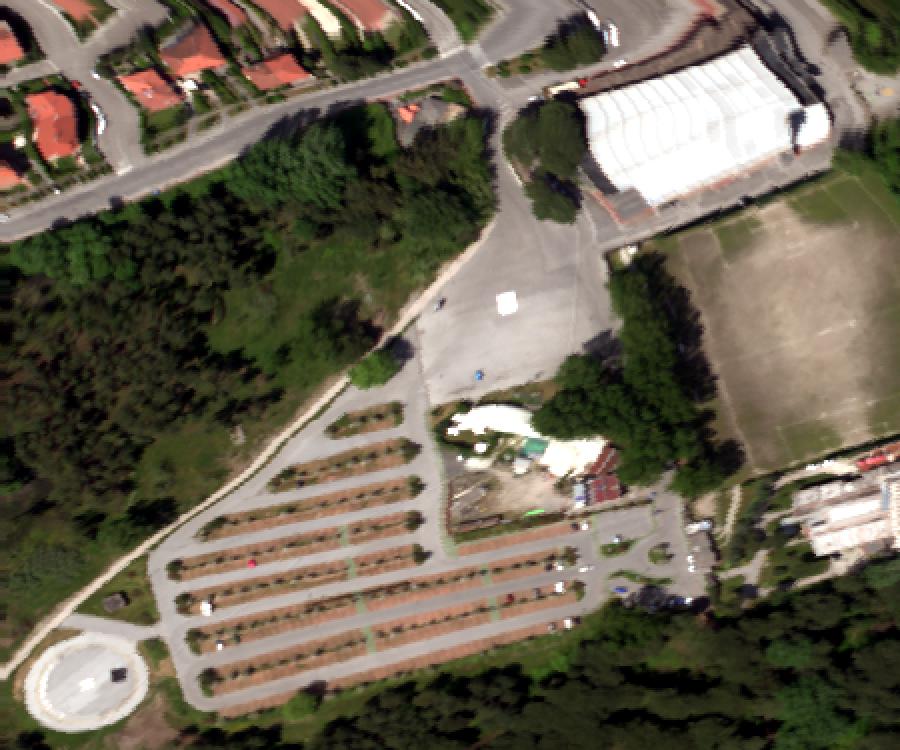}}
\subfigure[]{
\includegraphics[width=0.17\linewidth]{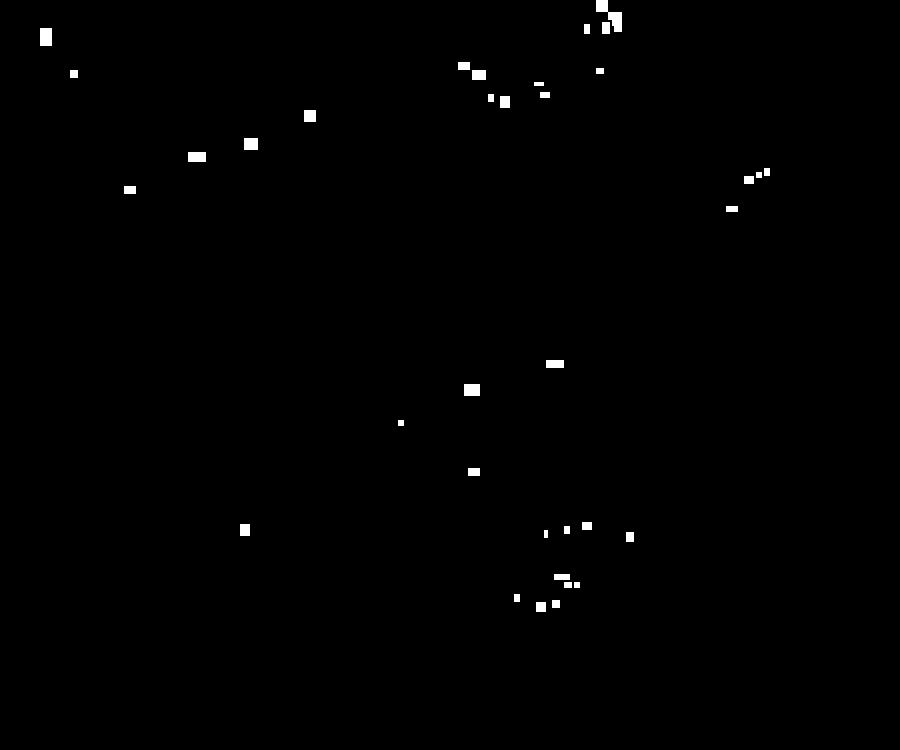}}
\subfigure[]{
\includegraphics[width=0.17\linewidth]{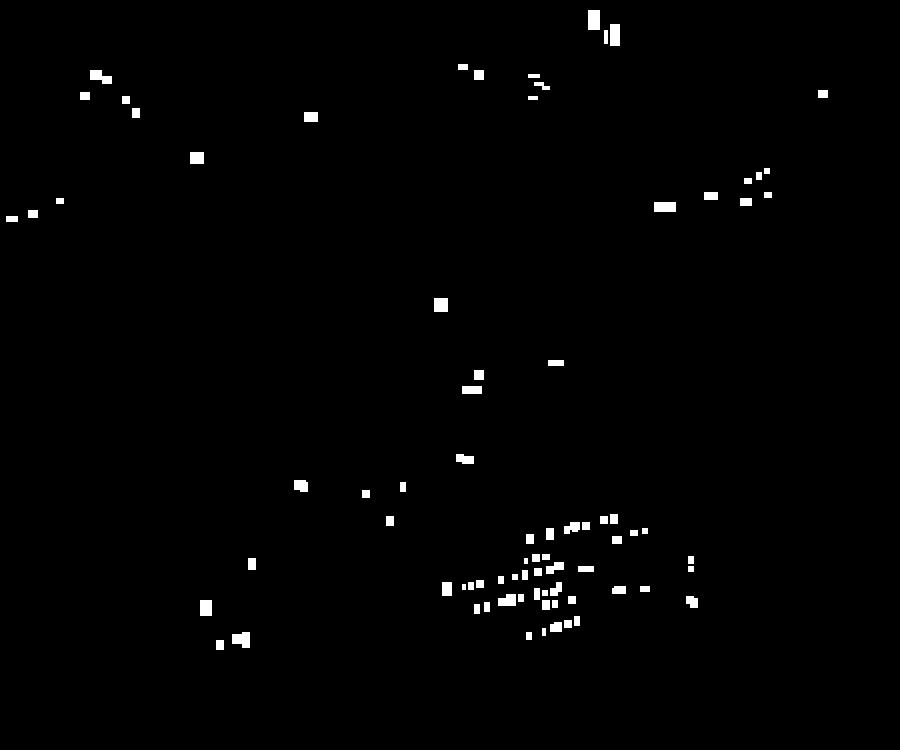}}
\caption{RGB-color images and the change mask of the Viareggio dataset: (a) D1F12H1, (b) D1F12H2, (c) D2F22H2, (d) change mask of ``D1F12H1$\_$D1F12H2", and (d) change mask of ``D1F12H1$\_$D2F22H2".}
\label{Viareggio}
\end{figure*}

{\bf 5) $\mathrm{W}^s$-subproblem}: With other variables being fixed, the
subproblem of updating $\mathrm{W}^s$ is:
\begin{equation*}
\mathrm{W}^{s\star}=\mathop{\arg\min}\limits_{\mathrm{W}^s}||\mathrm{W}^s||_{2,1}   +\frac{\mu}{2}||\mathrm{E}^s-\mathrm{W}^s+\frac{\mathrm{\bf Y}_3^s}{\mu}||_F^2.
\end{equation*}

Following Lemma 4.1 in \cite{liu2012robust}, the closed-form solution of the above function is:
\begin{equation*}
[\mathrm{W}^{s\star}]_{:,i}:=\left\{
\begin{aligned}
    \frac{||Q_{:,i}||_2-\frac{1}{\mu}}{||Q_{:,i}||_2}[Q_{:,i}], & \ \ \mathrm{if}||Q_{:,i}||_2>\frac{1}{\mu}; \\
    0, \ \ \ \ \ \ \ \ \ &\ \  \mathrm{otherwise},
\end{aligned}
\right.
\end{equation*}
where $Q=\mathrm{E}^s+\frac{\mathrm{\bf Y}_3^s}{\mu}$, and $Q_{:,i}$ denotes its $i$-th column.

{\bf 6) Updating the multipliers and $\mu$}:
\begin{equation*}
    \begin{split}
        \mathrm{\bf Y}_1^s&=\mathrm{\bf Y}_1^s+\mu(\mathrm{\bf X}^s-\mathrm{H}(\mathrm{C}+\mathrm{D}^s)-\mathrm{E}^s),\\
        \mathrm{\bf Y}_2^s&=\mathrm{\bf Y}_2^s+\mu(\textbf{1}^\top(\mathrm{C}+\mathrm{D}^s)-\textbf{1}^\top),\\
        \mathrm{\bf Y}_3^s&=\mathrm{\bf Y}_3^s+\mu(\mathrm{E}^s-\mathrm{W}^s),\\
        \mathrm{\bf Y}_4&=\mathrm{\bf Y}_4+\mu(\mathrm{C}-\mathrm{J}),\\
        \mu&=\min(\rho\mu,\mu_{\max}).
    \end{split}
\end{equation*}

The complete steps of the proposed SMSL model are shown in Algorithm \ref{alg1}, where the convergence conditions are:
\begin{equation*}
    \begin{split}
        &||\mathrm{\bf X}^s-\mathrm{H}(\mathrm{C}+\mathrm{D}^s)-\mathrm{E}^s||_\infty<\epsilon, \\
        &||\mathrm{E}^s-\mathrm{W}^s||_\infty<\epsilon, \\
        &||(\mathrm{C}+\mathrm{D}^s)^\top\textbf{1}-\mathbf{1}||_\infty<\epsilon, \\
        &\textrm{and}\ ||\mathrm{C}-\mathrm{J}||_\infty<\epsilon.
    \end{split}
\end{equation*}

\subsection{Decision Rule}
Through the aforementioned optimization process, the specific part of each pixel is derived and can be written as:
\begin{equation}
    \mathcal{S}_{\mathrm{\bf x}^s_i}=\mathrm{H}\mathrm{d}_i^s,
\end{equation}
where $\mathrm{x}_i^s$ is the $i$-th pixel of $\mathrm{\bf X}^s$ and $\mathrm{d}_i$ denotes the coefficient vector of the specific part. For bi-temporal images, the output of abnormal pixels is also influenced by the differences and the noise. Thus, we define the residual corresponding to $\mathrm{x}_i$ as the anomalous change detection result: 
\begin{equation}
    \mathcal{R}_{\mathrm{\bf x}_i}=||\mathcal{S}_{\mathrm{\bf x}^2_i}-\mathcal{S}_{\mathrm{\bf x}^1_i}||_2+||\mathrm{e}^2_i-\mathrm{e}^1_i||_2,
\end{equation}
where $\mathrm{e}_i^s$ denotes the $i$-th column of the noisy matrix $\mathrm{E}^s$. 

More generally, the decision rule can be written as the sum of the residuals:
\begin{equation}
    d(\mathrm{\bf x}_i)=\sum \mathcal{R}_{\mathrm{\bf x}_i}.
\end{equation}

\begin{figure}
\centering
\includegraphics[width=0.78\linewidth]{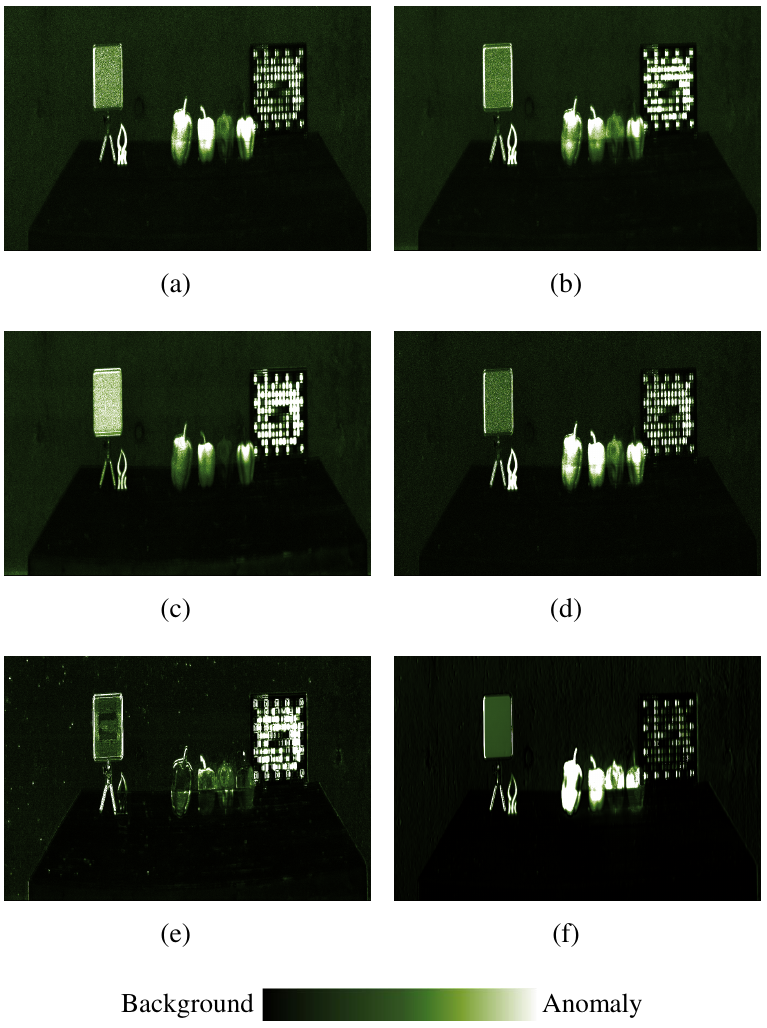}
\caption{The visualized anomalous change detection maps of Object$\_$1550$\_$1558 by (a) Difference RX, (b) CC, (c) CE, (d) SFA, (e) JSR, and (f) SMSL.}
\label{colormap_object}
\end{figure}

\begin{figure}
\centering
\includegraphics[width=0.78\linewidth]{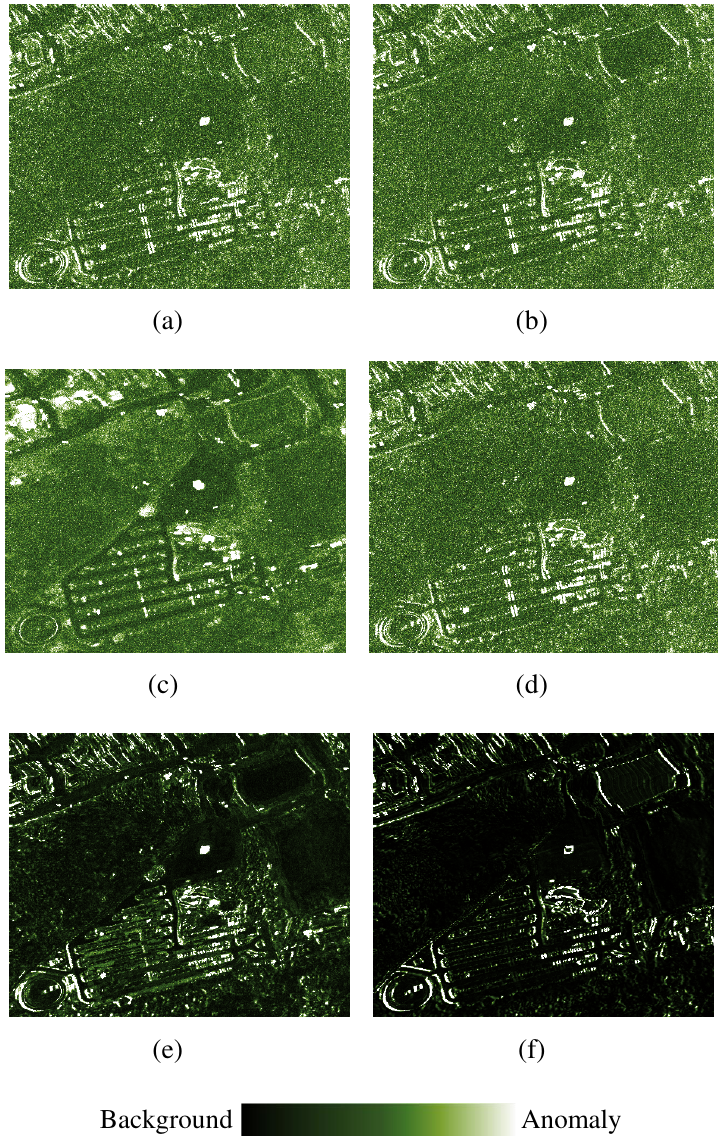}
\caption{The visualized anomalous change detection maps of ``D1F12H1$\_$D1F12H2" by (a) Difference RX, (b) CC, (c) CE, d) SFA, (e) JSR, and (f) SMSL.}
\label{colormap_via1}
\end{figure}
\subsection{Complexity and Convergence}
{\bf Complexity analysis.} The proposed model totally contains $3S+2$ optimization process, and the complexity is analyzed as follows. Considering that the size of the sketched dictionary is much smaller than the original inputs, the complexity of updating the coefficient matrix $\mathrm{C}$, $\mathrm{D}^s$ and the auxiliary variable $\mathrm{J}$ is simplified as $\mathcal{O}(N_\mathrm{H}^2N)$. The complexity of updating variable $\mathrm{E}^s$ and multiplier $\mathrm{Y}^s_1$ is $\mathcal{O}(LN_\mathrm{H}N)$ due to the matrix multiplication. And the complexity of the multiplier $\mathrm{Y}^s_2$ is $\mathcal{O}(N_\mathrm{H}N)$. For the subproblem $\mathrm{W}^s$, the complexity is $\mathcal{O}(LN)$. Then, the overall complexity of the SMSL model is $\mathcal{O}(SN_\mathrm{H}^2N+SLN_\mathrm{H}N+SN_\mathrm{H}N+SLN)$, which is basically $\mathcal{O}(SN_\mathrm{H}^2N+SLN_\mathrm{H}N)$. Since the size of the sketched dictionary and the image bands are much smaller than the image size, i.e., $N_\mathrm{H} \ll N$ and $L\ll N$, in contrast with traditional subspace learning methods \cite{zhang2018generalized, 9468429}, the computational consumption of our model is obviously decreased and the limitations of multi-view learning are greatly addressed.

{\bf Convergence analysis.} Unfortunately, the convergence of the ALM method with three or more pending matrices is very difficult to prove theoretically. Inspired by previous research\cite{zhang2018generalized, 9468429, zheng2020feature}, the gap generated in each iteration is calculated and shown in the experimental results. We find that the SMSL model can be expected to have good convergence properties.

\section{Experiments}
\begin{figure}
\centering
\includegraphics[width=0.78\linewidth]{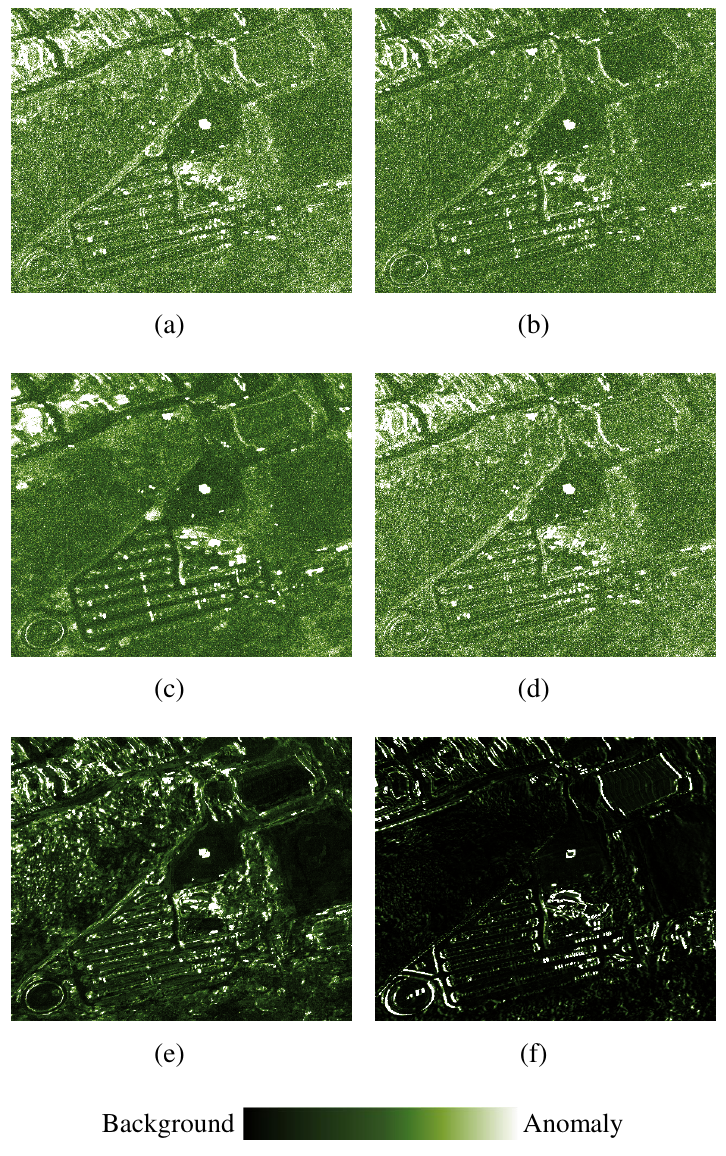}
\caption{The visualized anomalous change detection maps of ``D1F12H1$\_$D2F22H2" by (a) Difference RX, (b) CC, (c) CE, (d) SFA, (e) JSR, and (f) SMSL.}
\label{colormap_via2}
\end{figure}
In this section, experiments are conducted on a natural HSI dataset and two hyperspectral remote sensing image pairs for bi-temporal anomalous change detection. Accordingly, the results are shown with corresponding analysis and the proposed method is compared with five state-of-the-art anomalous change detectors: the difference RX \cite{zhou2016novel}, CC \cite{schaum2015theoretical}, CE \cite{schaum2004hyperspectral}, SFA \cite{wu2015hyperspectral}, and JSR \cite{wu2018hyperspectral}. The convergence and the parametric sensitivity are tested as well. All the experiments are
conducted in Matlab on an Intel Core i7-8550U
CPU with 16 GB of RAM.
\begin{figure*}[h]
\centering
\subfigure[]{
\includegraphics[width=0.29\linewidth]{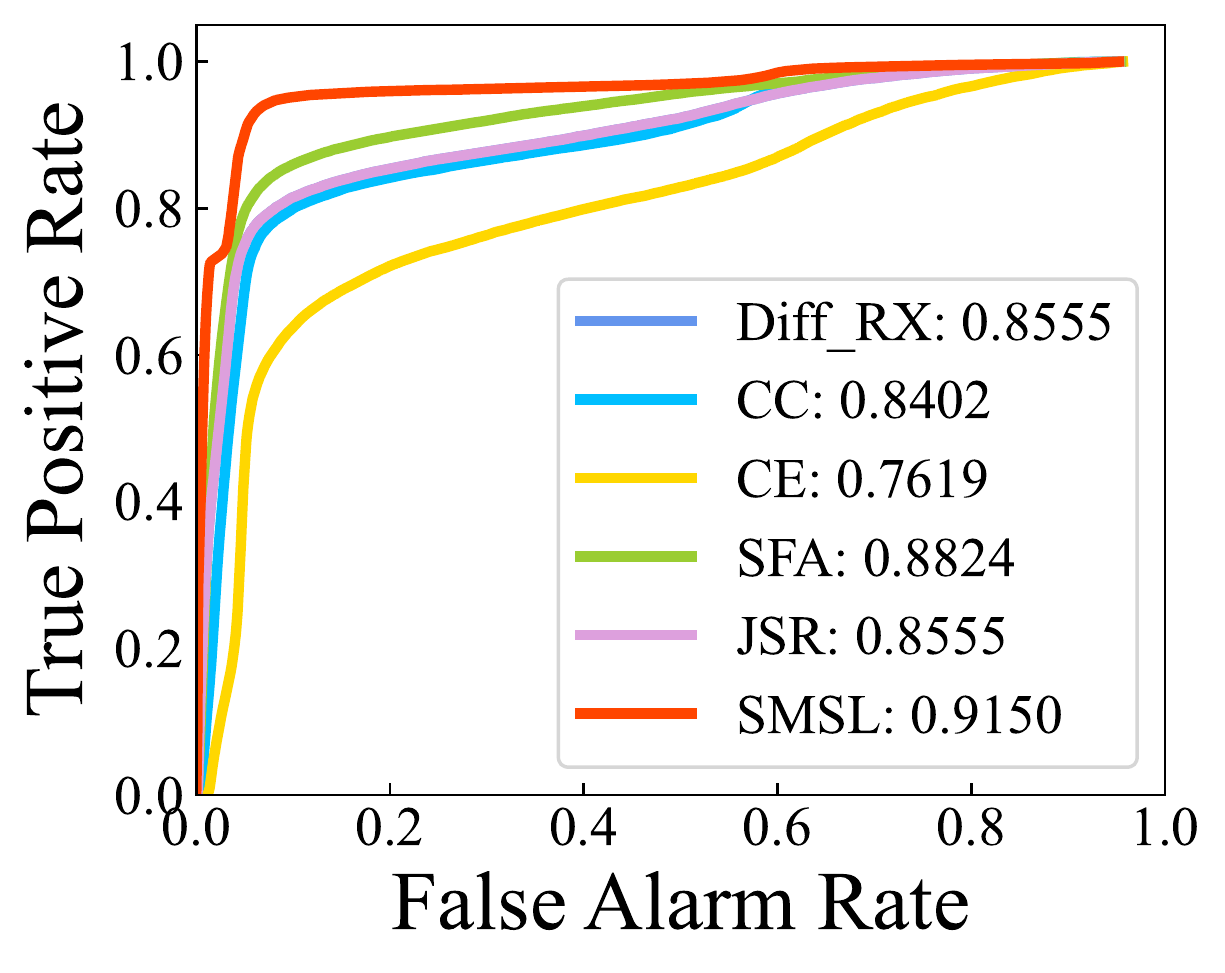}}
\subfigure[]{
\includegraphics[width=0.29\linewidth]{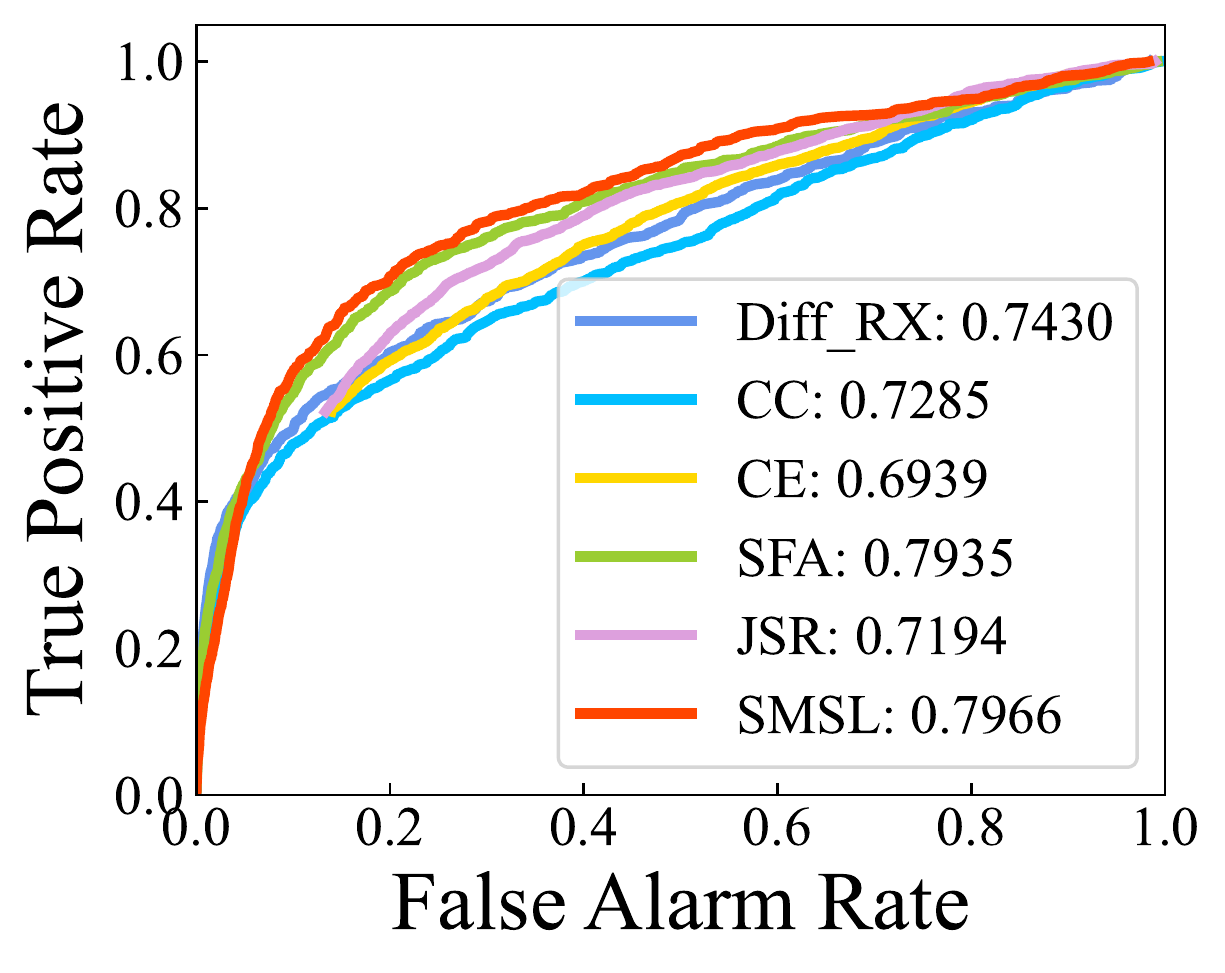}}
\subfigure[]{
\includegraphics[width=0.29\linewidth]{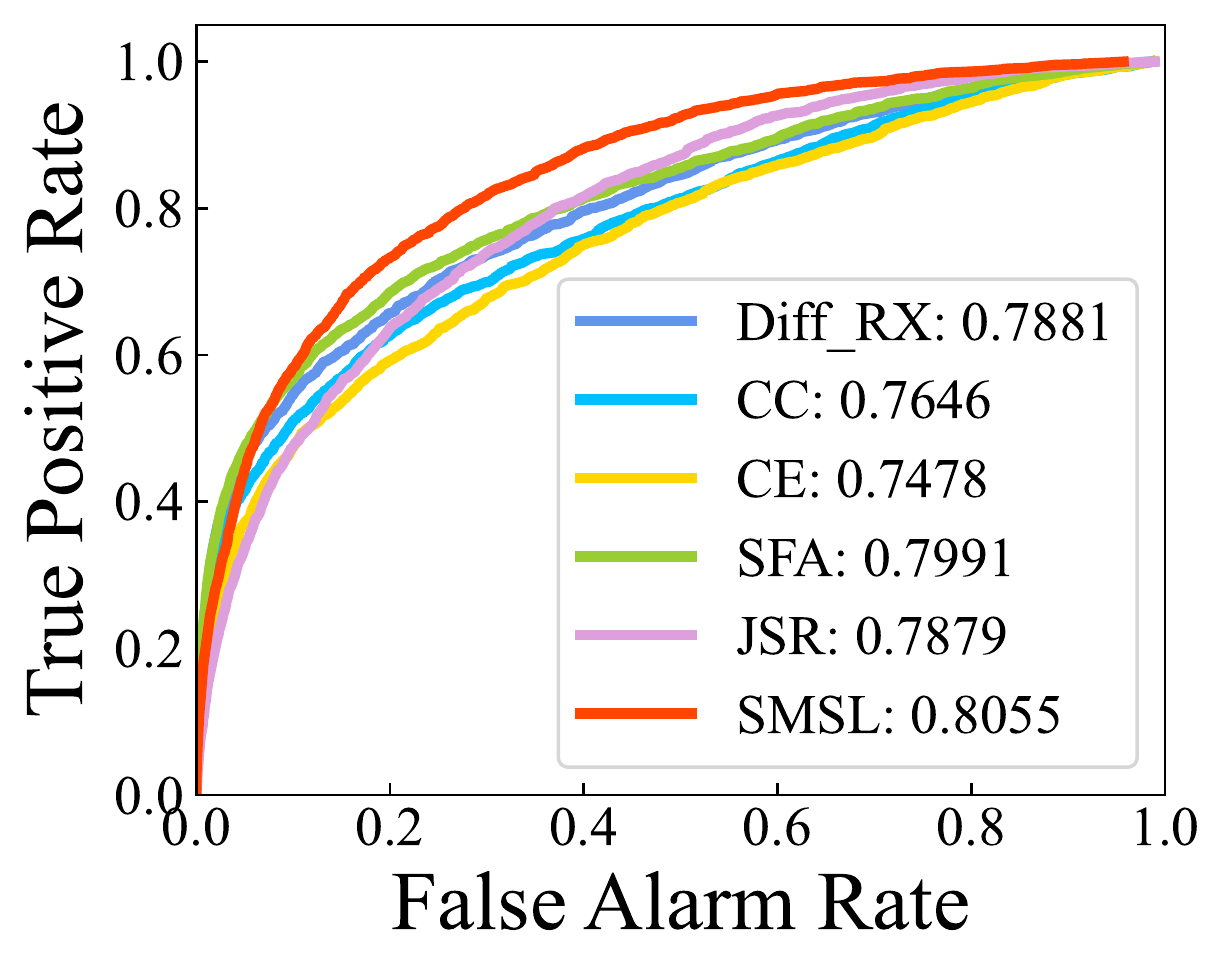}}
\caption{The ROC curves of all methods in (a) Object$\_$1550$\_$1558, (b) D1F12H1$\_$D1F12H2, and (c) D1F12H1$\_$D2F22H2.}
\label{ROC}
\end{figure*}
\subsection{HSI Datasets}
\subsubsection{Object$\_$1550$\_$1558 dataset} This dataset was originally derived from the BGU iCVL hyperspectral image dataset\footnote{http://icvl.cs.bgu.ac.il/hyperspectral/}. A Specim PS Kappa D$\times$4 hyperspectral camera and a rotary stage for spatial scanning are used to acquire the data \cite{arad_and_ben_shahar_2016_ECCV}. We selected one image pair from the database and cropped a sub-image for anomalous change detection. Each sub-image contains 600$\times$900 spatial resolution over 31 spectral bands from 400 nm to 700 nm at 10 nm increments. The change mask was created in ENVI by combining the human observation and the pre-detection result of the differential image. Figs. \ref{object15501558}(a) and (b) are the RGB-colored images of the \textit{Object$\_$1550} and the \textit{Object$\_$1558}, respectively, and Fig. \ref{object15501558}(c) is the change mask of the changed objects.
\subsubsection{Viareggio Datasets}
The Viareggio dataset \cite{7430258} provided by the Viareggio 2013 Trial contains three hyperspectral remote sensing images of the same study area. Both D1F12H1 and D1F12H2 were acquired on May 8, 2013, where the illumination conditions are very similar. D2F22H2 was acquired the following day, on May 9, 2013, when the illumination conditions were quite different. For anomalous change detection, this dataset is usually divided into two pairs: ``D1F12H1$\_$D1F12H2'' and ``D1F12H1$\_$D2F22H2''. Each image contains 375$\times$450 pixels with 128 spectral bands. The ground-truth mask for pair-wise change detection is provided by the trial. Fig. \ref{Viareggio} illustrates the RGB-colored images and the change masks of two pairs.
\subsection{Experimental Results}
\subsubsection{Performance Comparison}
Since the sketched dictionary is constructed by the JLT transformation, the result could be influenced by the random matrix. Taking into consideration the computation consumption of the multi-view subspace learning model, we repeat our construction process 10 times, and the result is derived using the average sketched dictionary. To quantitatively evaluate the performances of different methods, the receiver operating characteristic curve (ROC) and the area under the ROC curve (AUC) values are utilized for illustration. For a fair comparison, we tuned the parameters of the compared methods and reported the best results.

The visualized results of three datasets are shown in Figs. \ref{colormap_object}$-$\ref{colormap_via2}, and the ROC curves and corresponding AUC values of all the algorithms are shown in Fig. \ref{ROC}. It can be seen that for the natural hyperspectral dataset, the mirror and the palette located on the table are quite different from the surroundings, but they are not changed in different scenes. The compared methods failed to suppress the mirror and the palette, and have relatively high detection output. In contrast, the proposed method is better able to identify the changed objects from unusual backgrounds, while the palette and the mirror have relatively small outputs. For the ``D1F12H1$\_$D1F12H2'' and ``D1F12H1$\_$D2F22H2'' image pairs, the background materials are quite complicated, while the anomalous pixels are rare and small, which means highlighting the anomalous changes from the background should be more difficult. It can be seen from Figs. \ref{colormap_via1} and \ref{colormap_via2} that the difference RX, CC, CE, and SFA methods failed to suppress the background pixels, and result in bad visualization detection maps. Compared to other methods, the JSR and SMSL methods are more robust to the backgrounds. Combining the quantitative evaluations shown in Fig. \ref{ROC}, we can clearly see that the ROC curves of the proposed SMSL is more close to the top left corner and reaches the largest AUC values in all three datasets.

\begin{figure*}
\centering
\subfigure[Five pixels randomly selected from anomalous changes]{
\includegraphics[width=0.97\linewidth]{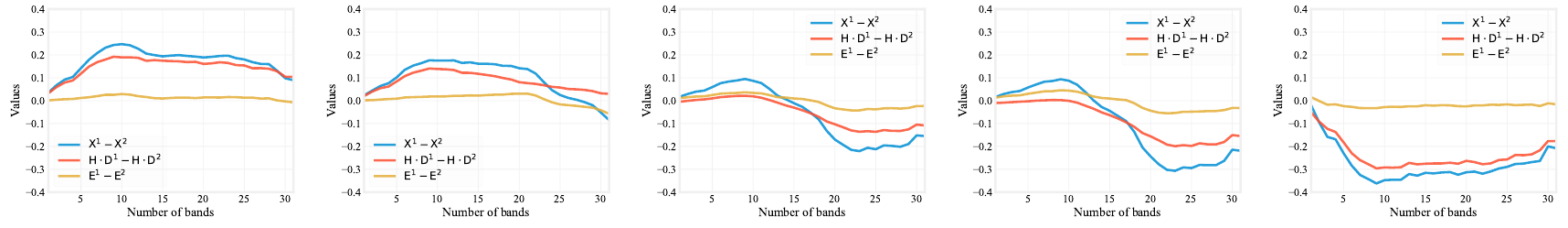}}

\subfigure[Five pixels randomly selected from backgrounds]{
\includegraphics[width=0.97\linewidth]{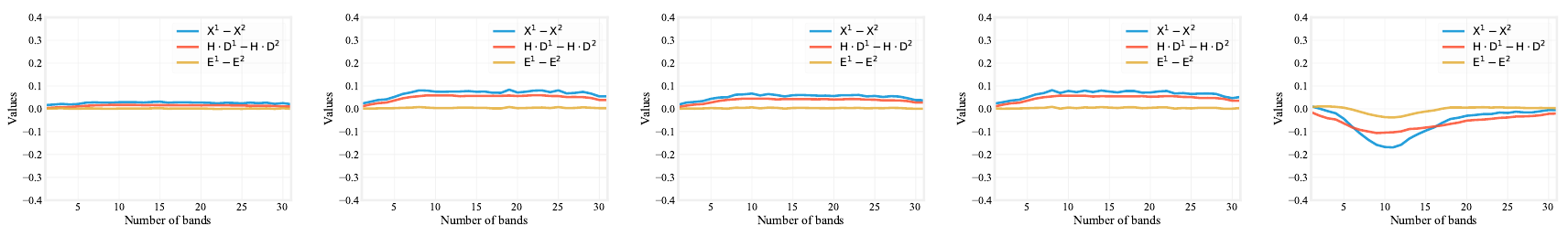}}

\subfigure[Five pixels randomly selected from anomalous changes]{
\includegraphics[width=0.97\linewidth]{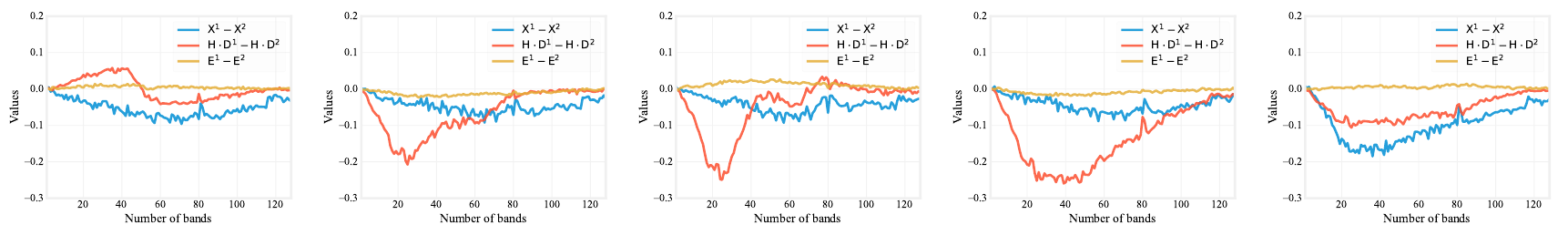}}

\subfigure[Five pixels randomly selected from backgrounds]{
\includegraphics[width=0.97\linewidth]{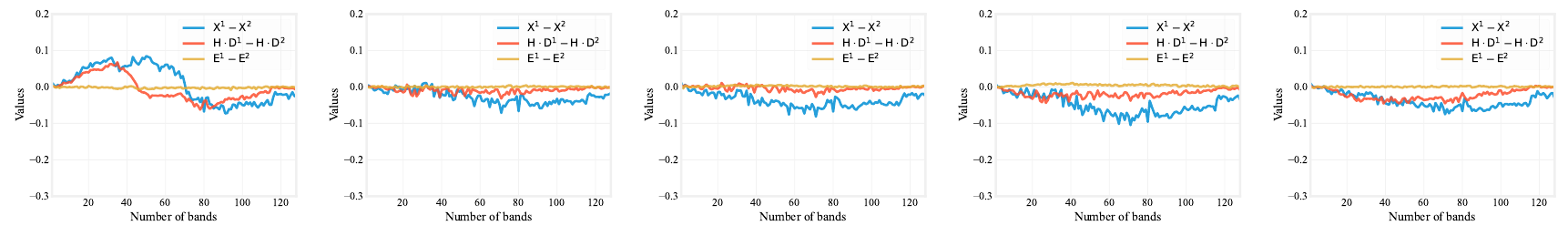}}

\subfigure[Five pixels randomly selected from anomalous changes]{
\includegraphics[width=0.97\linewidth]{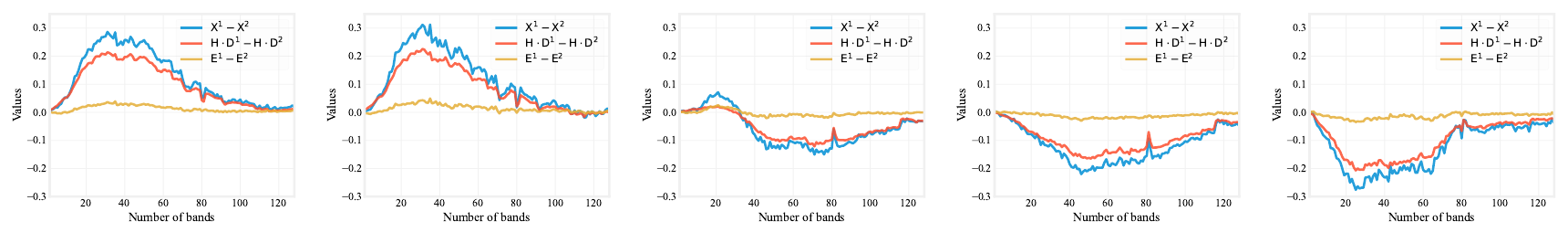}}

\subfigure[Five pixels randomly selected from backgrounds]{
\includegraphics[width=0.97\linewidth]{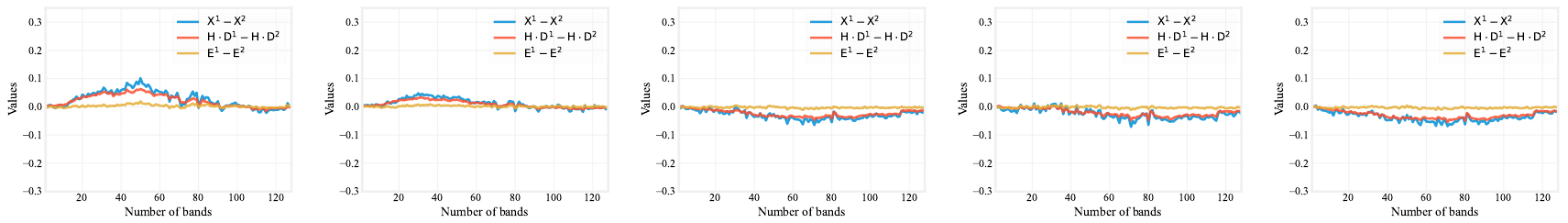}}
\caption{Spectral curves of the randomly selected pixels in three datasets: (a)$-$(b) Object$\_$1550$\_$1558, (b)$-$(c) D1F12H1$\_$D1F12H2, and (d)$-$(e) D1F12H1$\_$D2F22H2.}
\label{spectral}
\end{figure*}

To better illustrate the effectiveness of the proposed model, we randomly select five pixels that belong to the anomalous changes and the backgrounds from the images and make a comparison of their spectral characteristics for the three datasets. The spectral difference between the original inputs $\mathrm{\bf X}^1$ and $\mathrm{\bf X}^2$, the specific matrix $\mathrm{H}\cdot\mathrm{D}^1$ and $\mathrm{H}\cdot\mathrm{D}^2$, and the noisy matrix $\mathrm{E}^1$ and $\mathrm{E}^2$, are given in Fig. \ref{spectral}. We can see that, overall the anomalous changes have relatively larger spectral differences between the original inputs than the backgrounds. For the ``Object$\_$1550$\_$1558'' and the ``D1F12H1$\_$D2F22H2'' datasets, the differences between the two times are well preserved on anomalous pixels, meanwhile, the differences between background pixels are also well suppressed. And for the ``D1F12H1$\_$D1F12H2'' dataset, the SMSL model can enlarge the differences between anomalous pixels and decrease the differences between background pixels.

\subsubsection{Parametric Analysis}
\begin{figure}[h]
\centering
\includegraphics[width=0.75\linewidth]{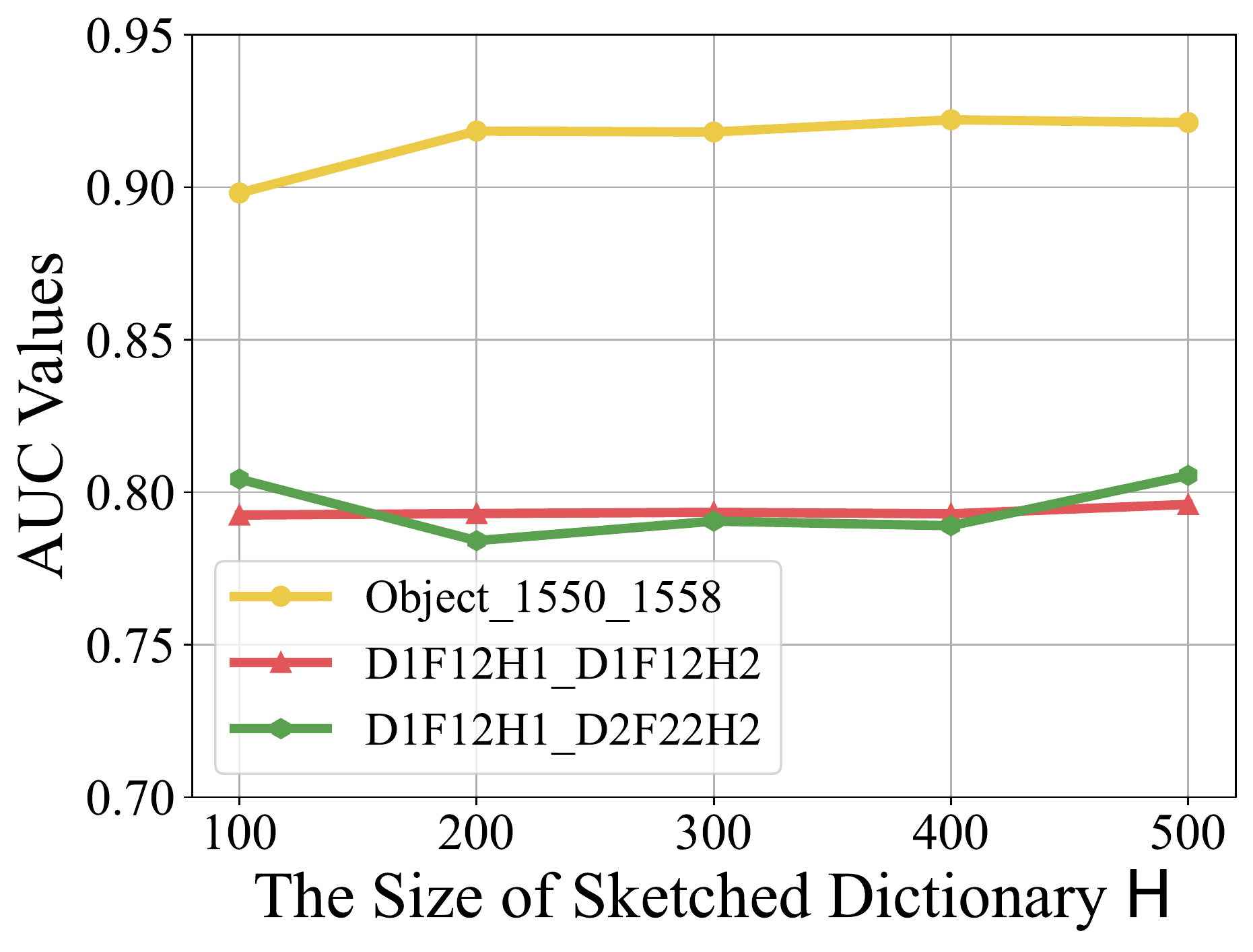}
\caption{The AUC values of the SMSL model with respect to the size of sketched dictionary in three datasets.}
\label{sketched}
\end{figure}

As has been discussed before, the sketched dictionary we designed aims to preserve the most important information while effectively saving computational consumption and storage. However, it is clear that with the reduction of the size of the sketching matrix, the consumption of the optimization process is saved, meanwhile, we may lose more information that is originally contained in the inputs. Therefore, we adjust the size of sketched dictionary from 100 to 500, and test the performances of SMSL in three datasets. As shown in Fig. \ref{sketched}, the AUC values of the proposed SMSL model are generally improved as the size of sketched dictionary increases. To find a compromise between preserving the majority information and effectively computation, the size of $\mathrm{H}$ is set to 500.

\begin{figure*}
\centering
\subfigure[$\lambda_1=0.1$]{
\includegraphics[width=0.18\linewidth]{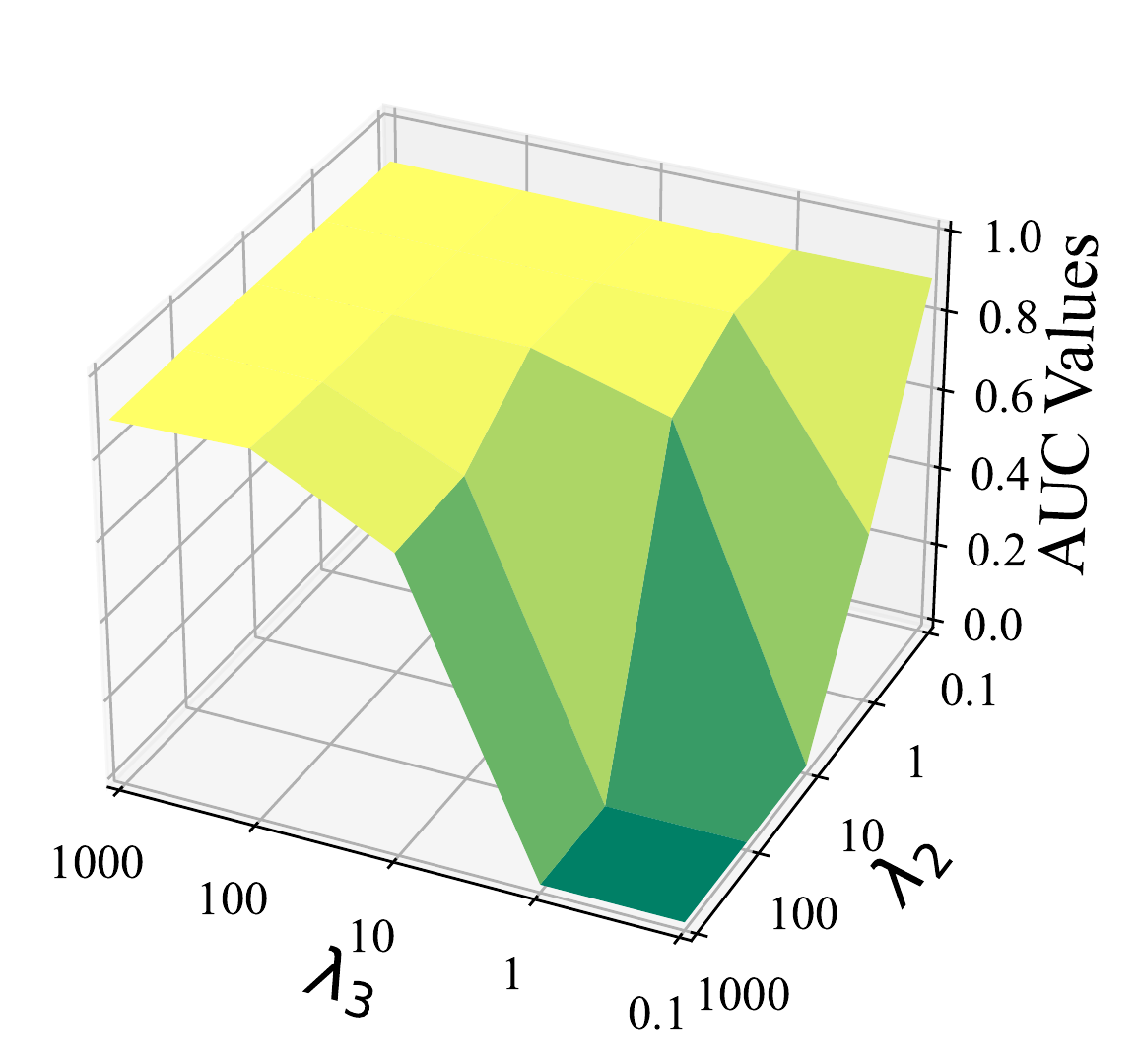}}
\subfigure[$\lambda_1=1$]{
\includegraphics[width=0.18\linewidth]{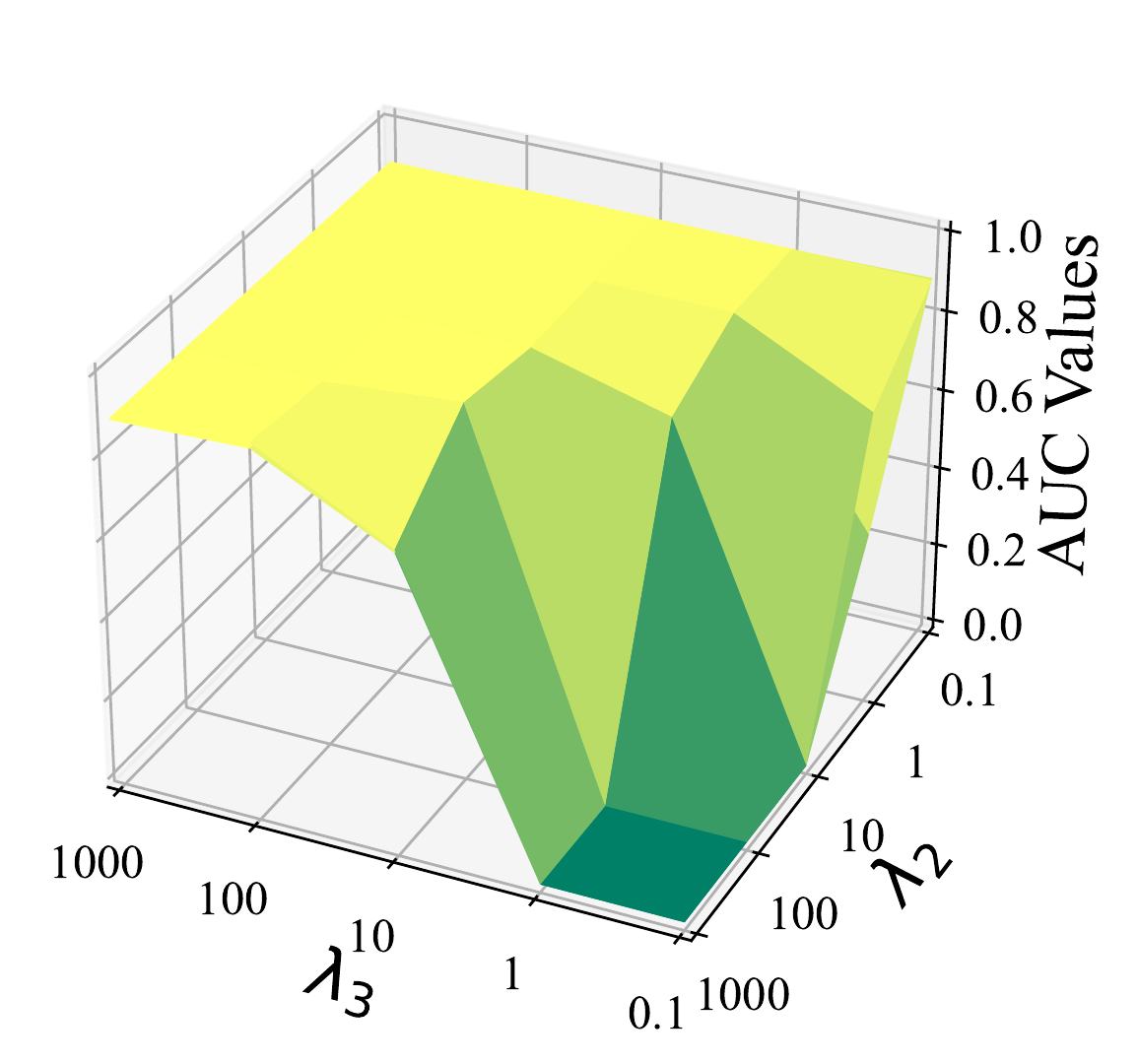}}
\subfigure[$\lambda_1=10$]{
\includegraphics[width=0.18\linewidth]{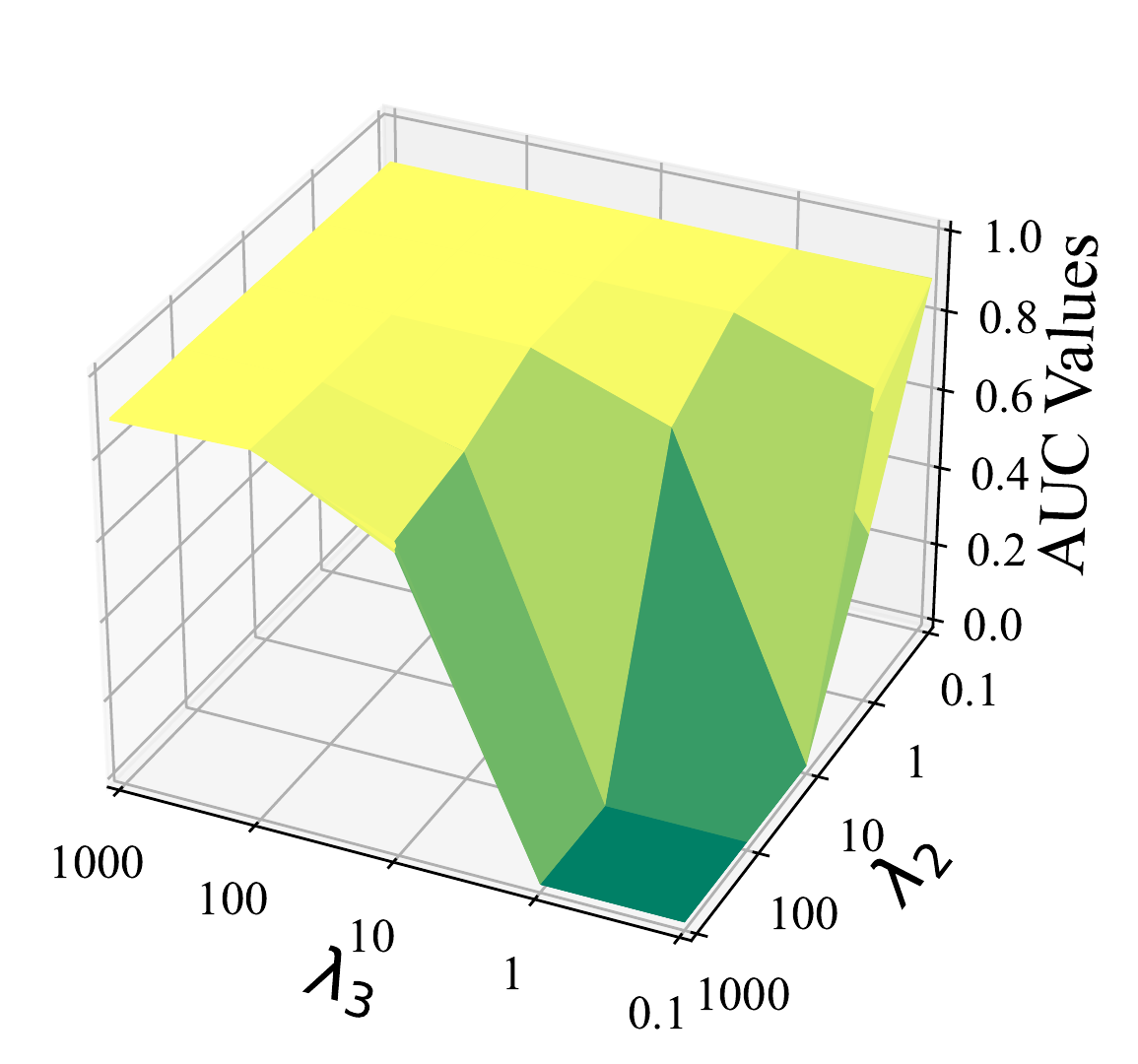}}
\subfigure[$\lambda_1=100$]{
\includegraphics[width=0.18\linewidth]{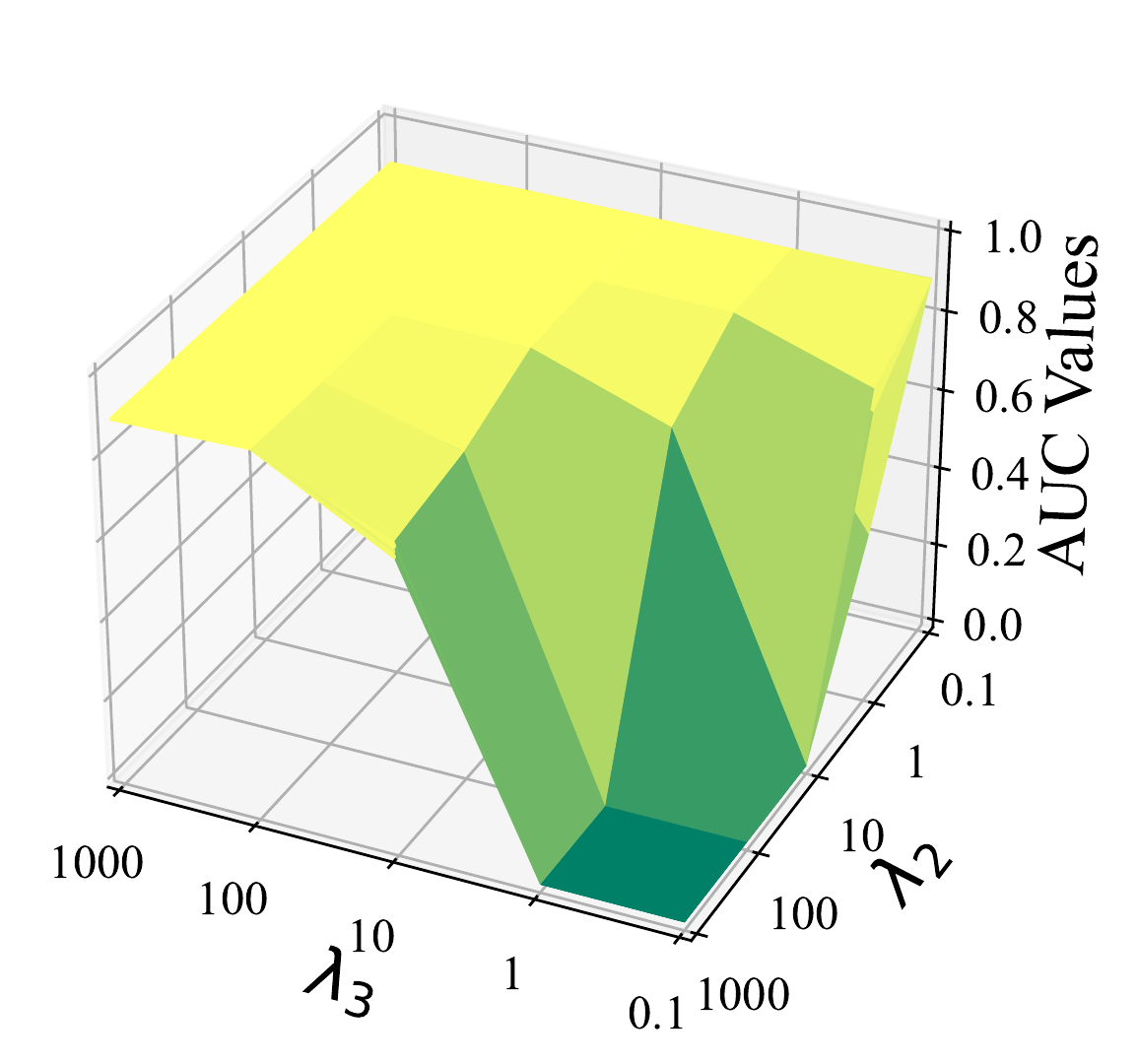}}
\subfigure[$\lambda_1=1000$]{
\includegraphics[width=0.18\linewidth]{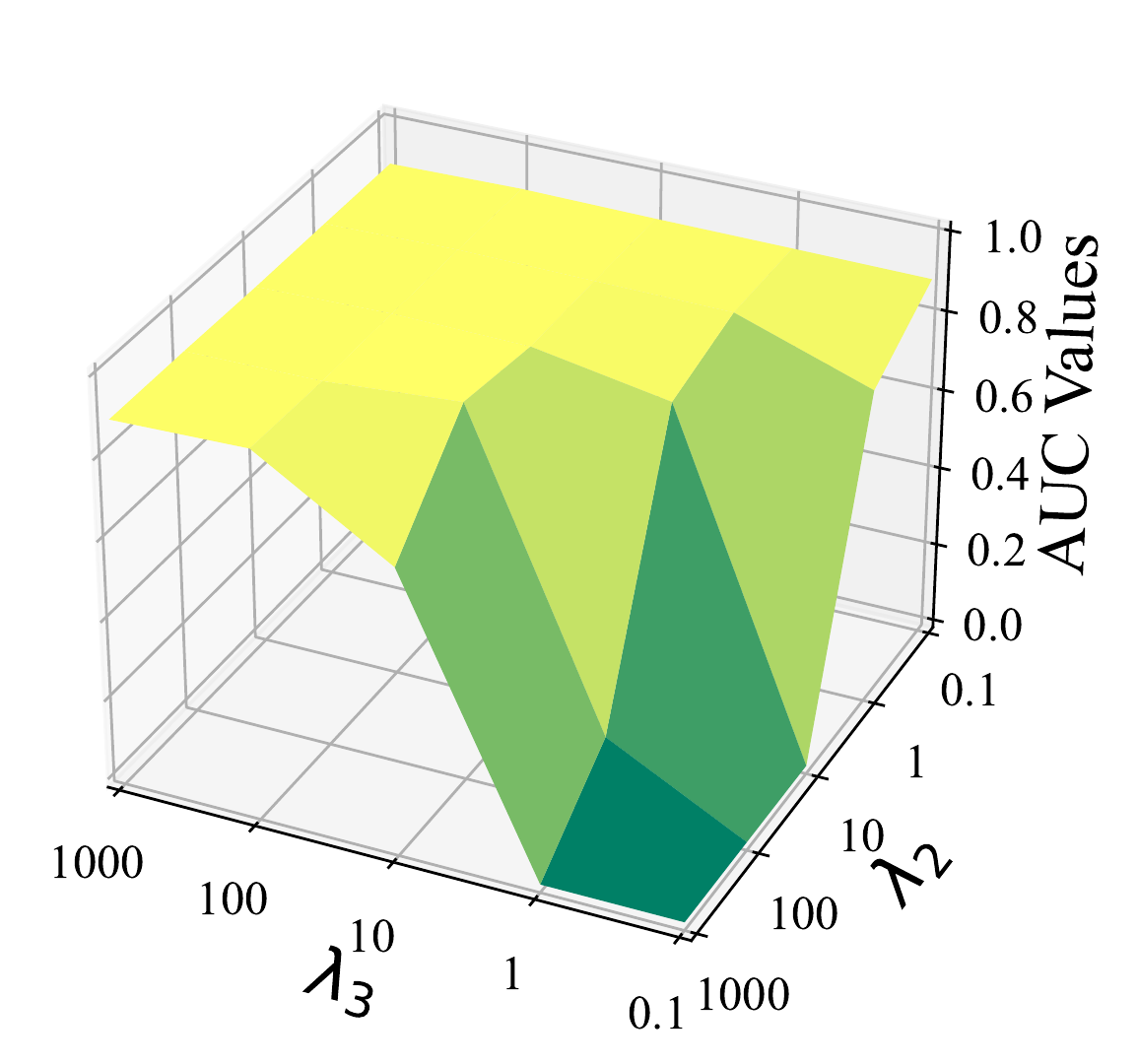}}

\subfigure[$\lambda_1=0.1$]{
\includegraphics[width=0.18\linewidth]{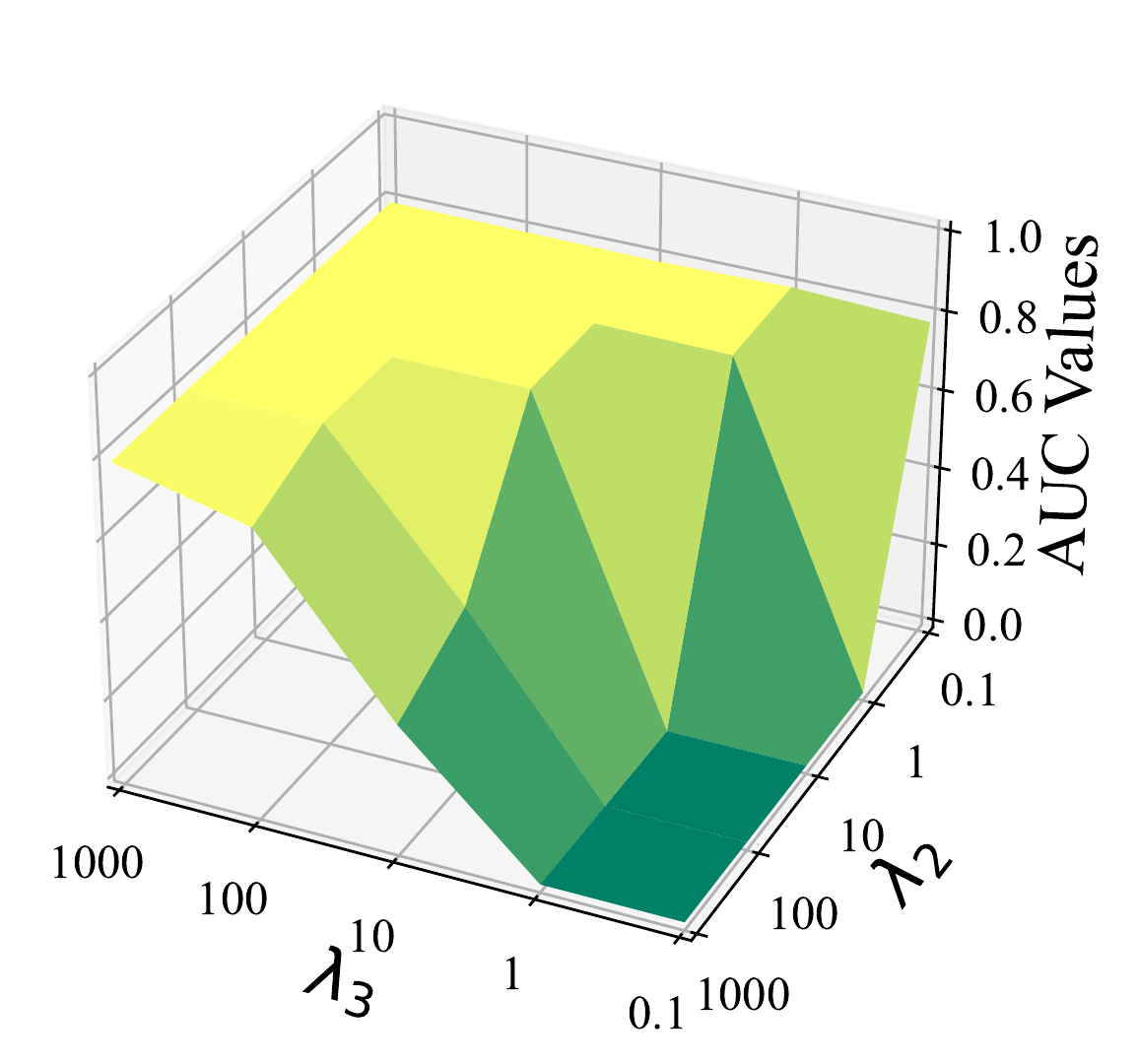}}
\subfigure[$\lambda_1=1$]{
\includegraphics[width=0.18\linewidth]{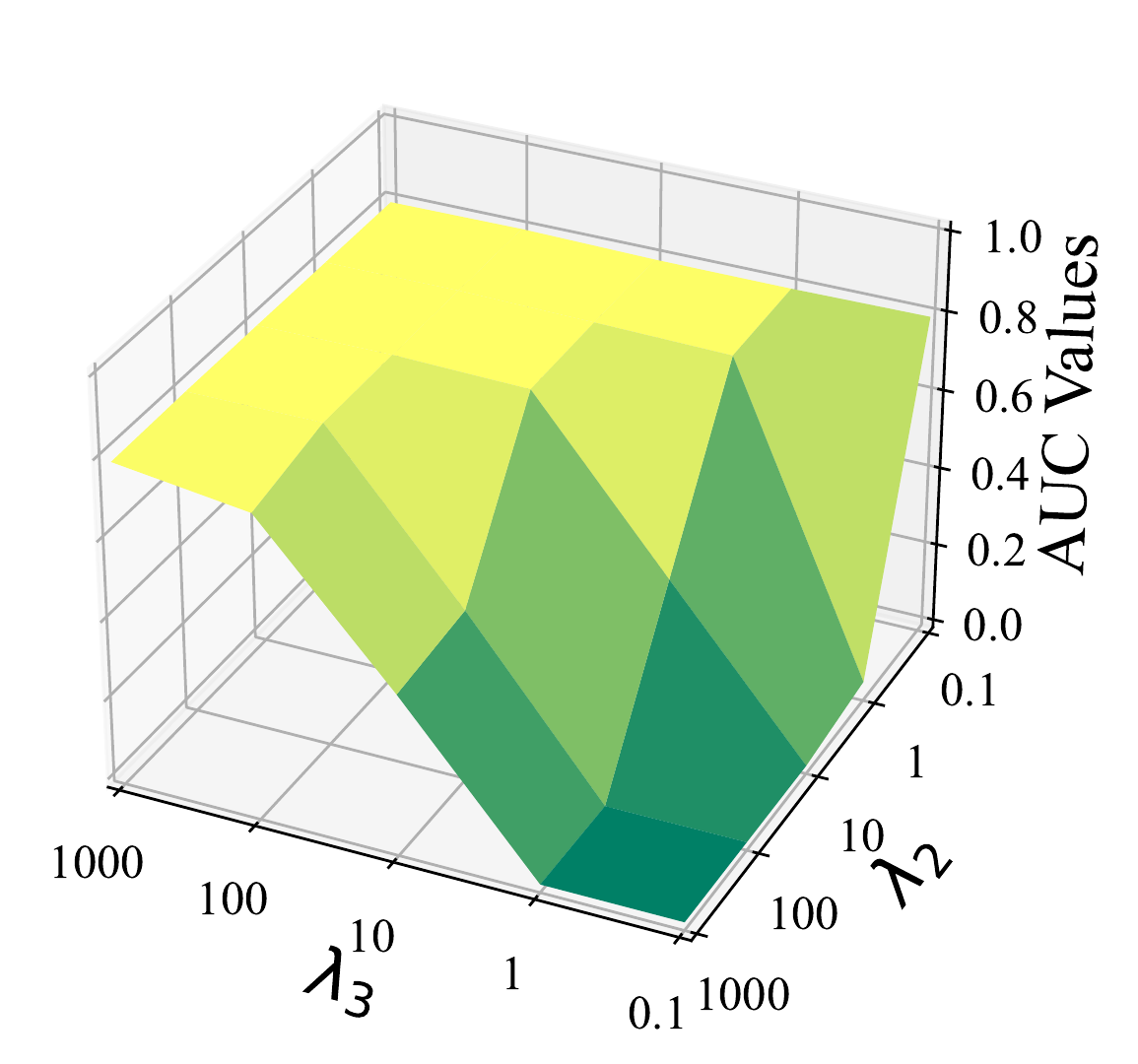}}
\subfigure[$\lambda_1=10$]{
\includegraphics[width=0.18\linewidth]{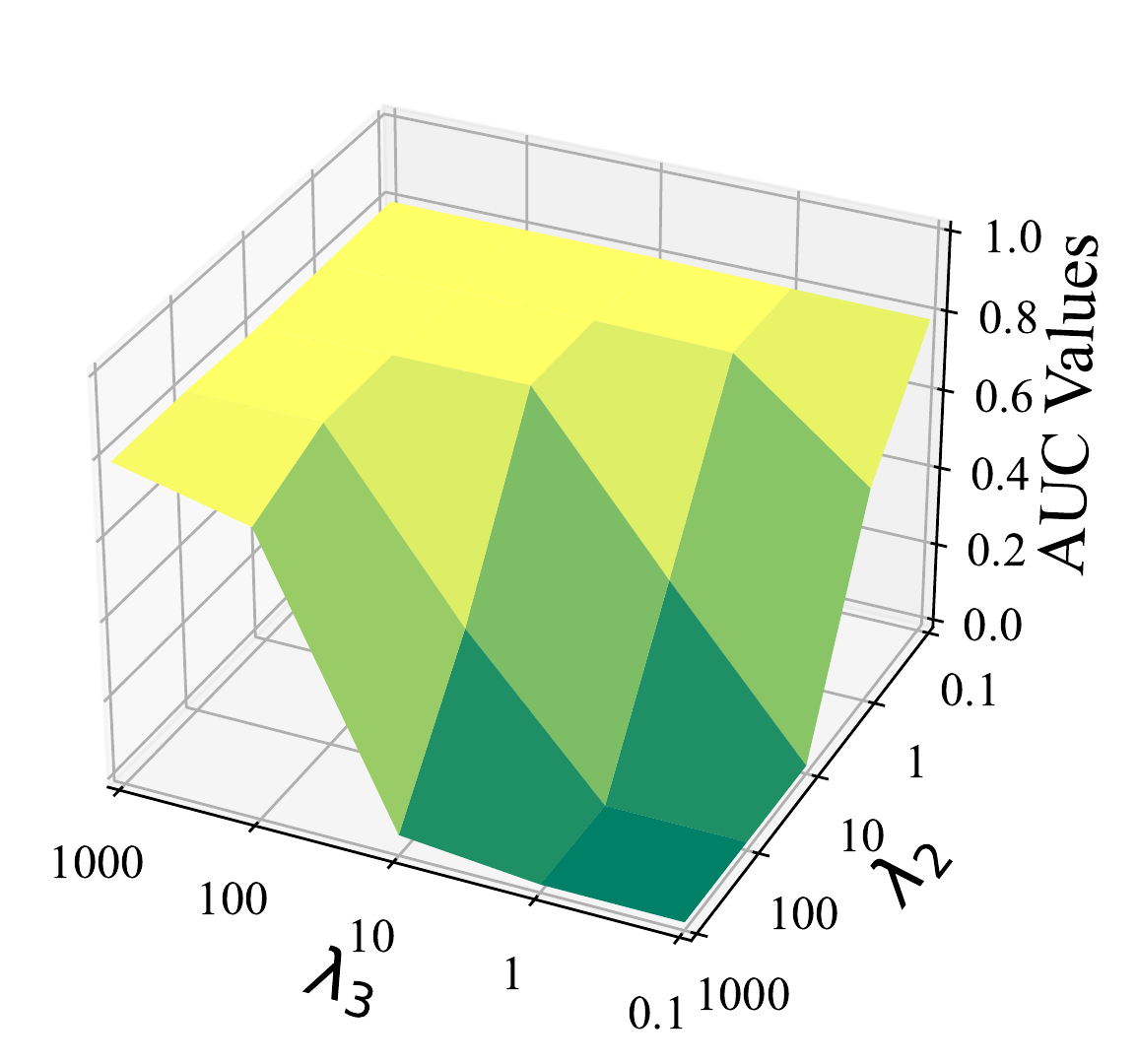}}
\subfigure[$\lambda_1=100$]{
\includegraphics[width=0.18\linewidth]{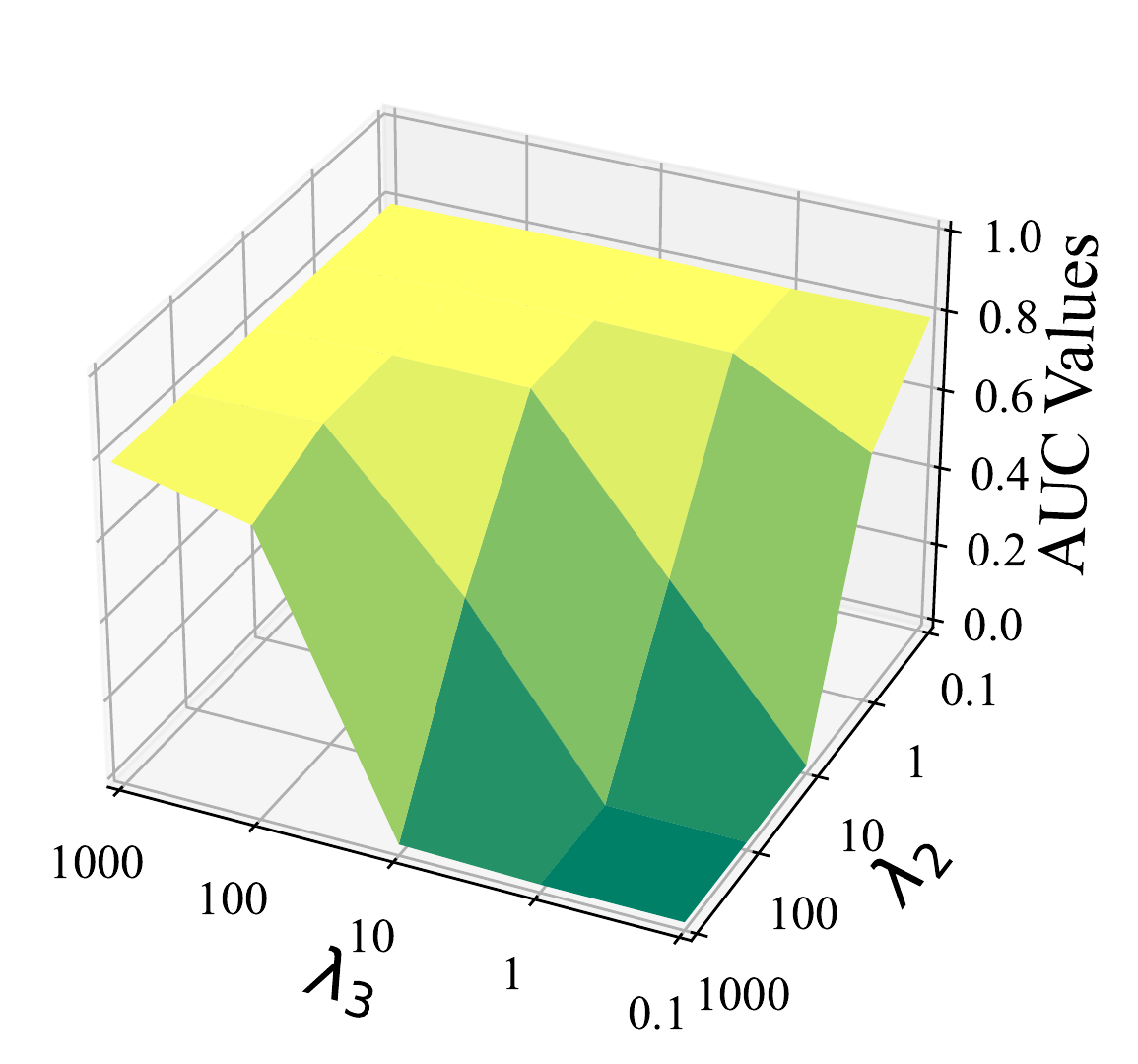}}
\subfigure[$\lambda_1=1000$]{
\includegraphics[width=0.18\linewidth]{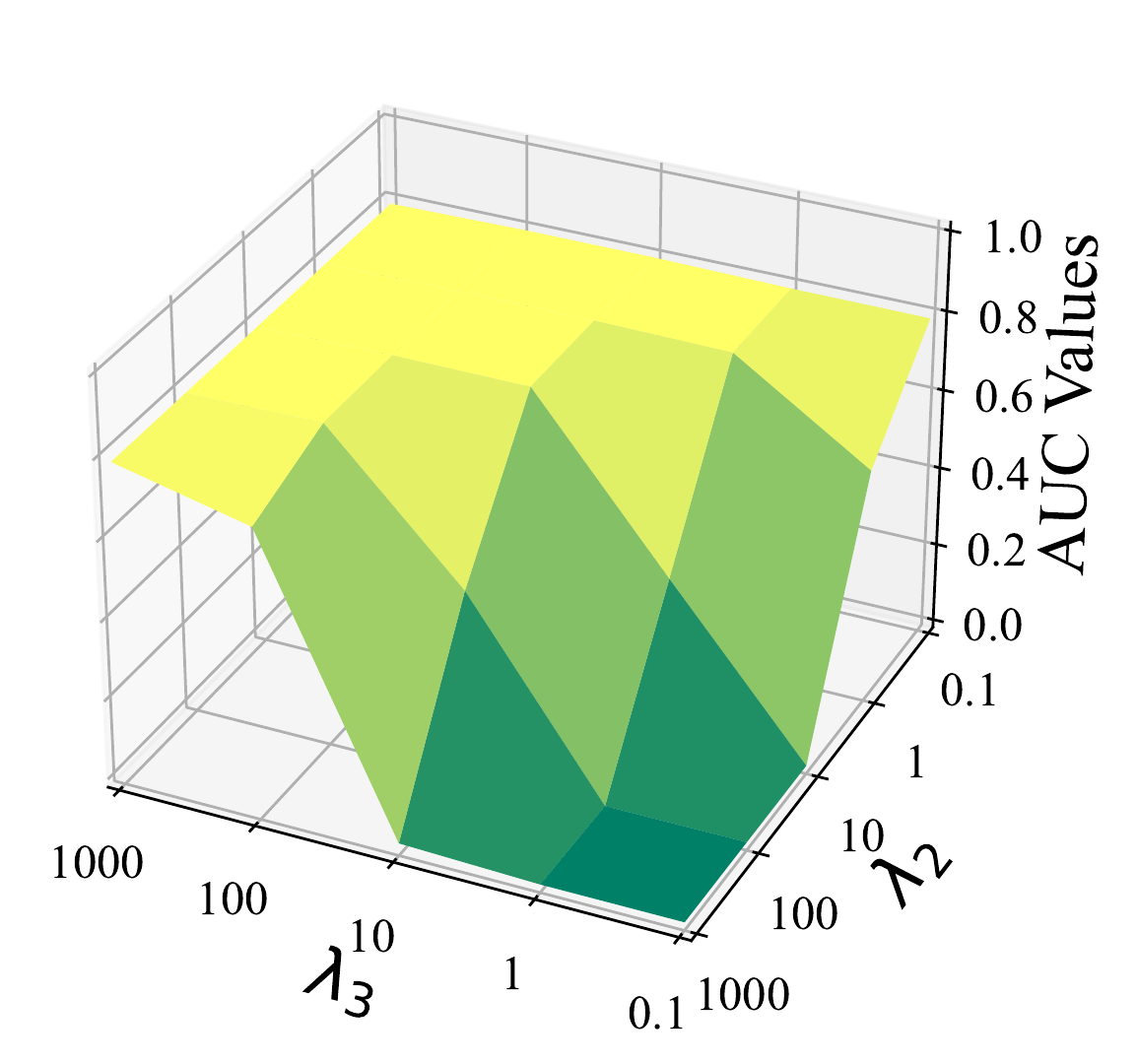}}

\subfigure[$\lambda_1=0.1$]{
\includegraphics[width=0.18\linewidth]{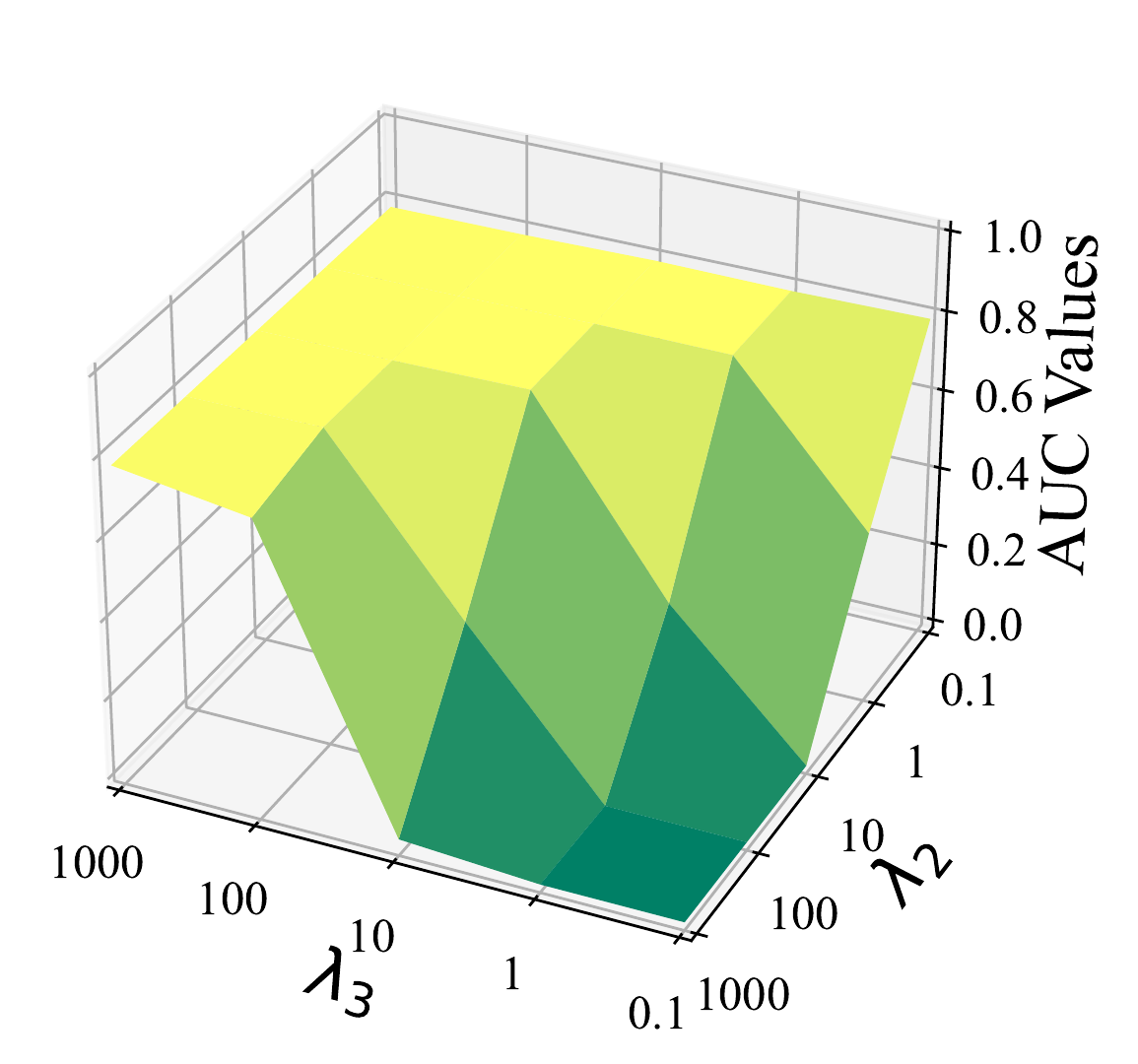}}
\subfigure[$\lambda_1=1$]{
\includegraphics[width=0.18\linewidth]{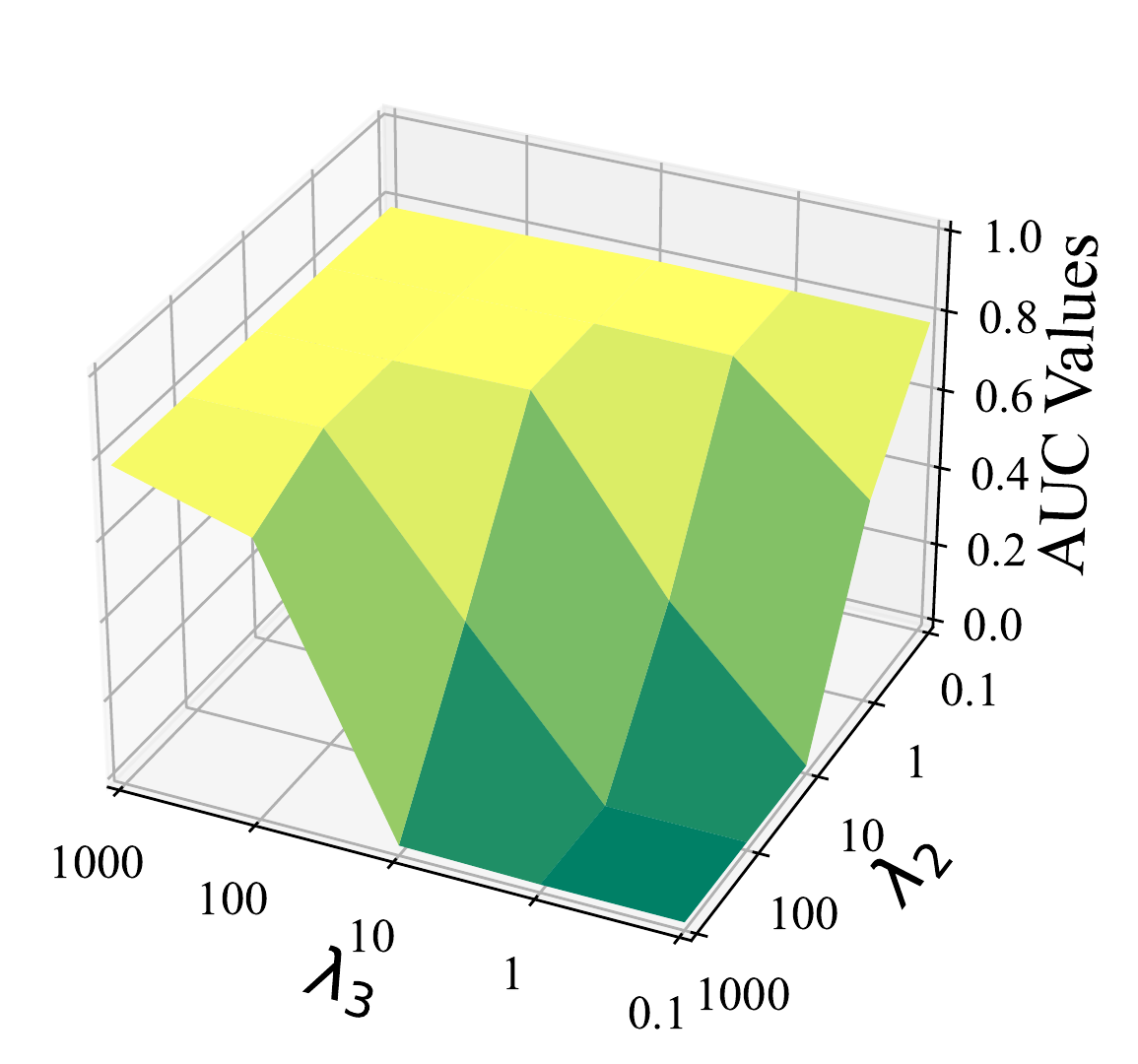}}
\subfigure[$\lambda_1=10$]{
\includegraphics[width=0.18\linewidth]{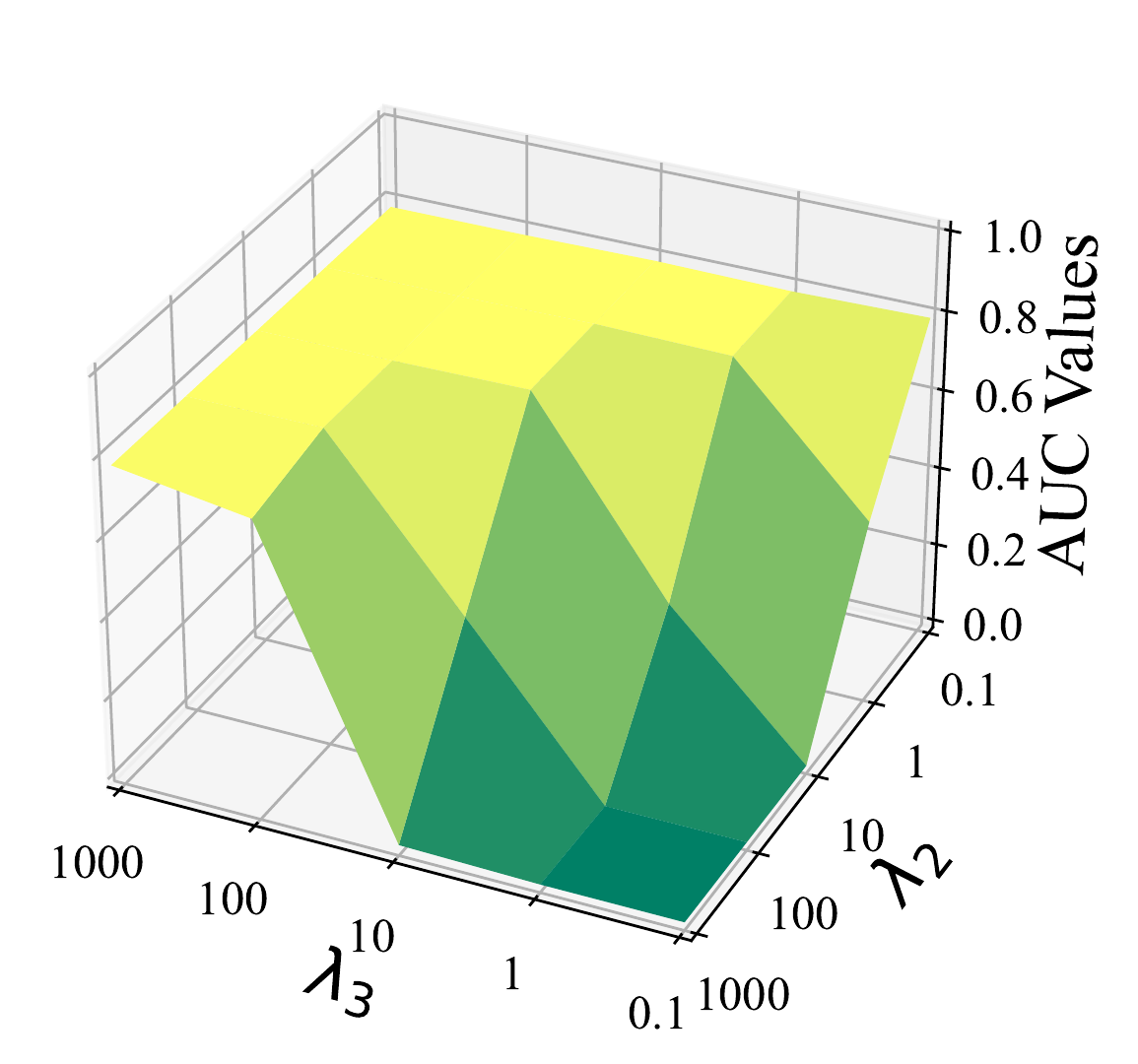}}
\subfigure[$\lambda_1=100$]{
\includegraphics[width=0.18\linewidth]{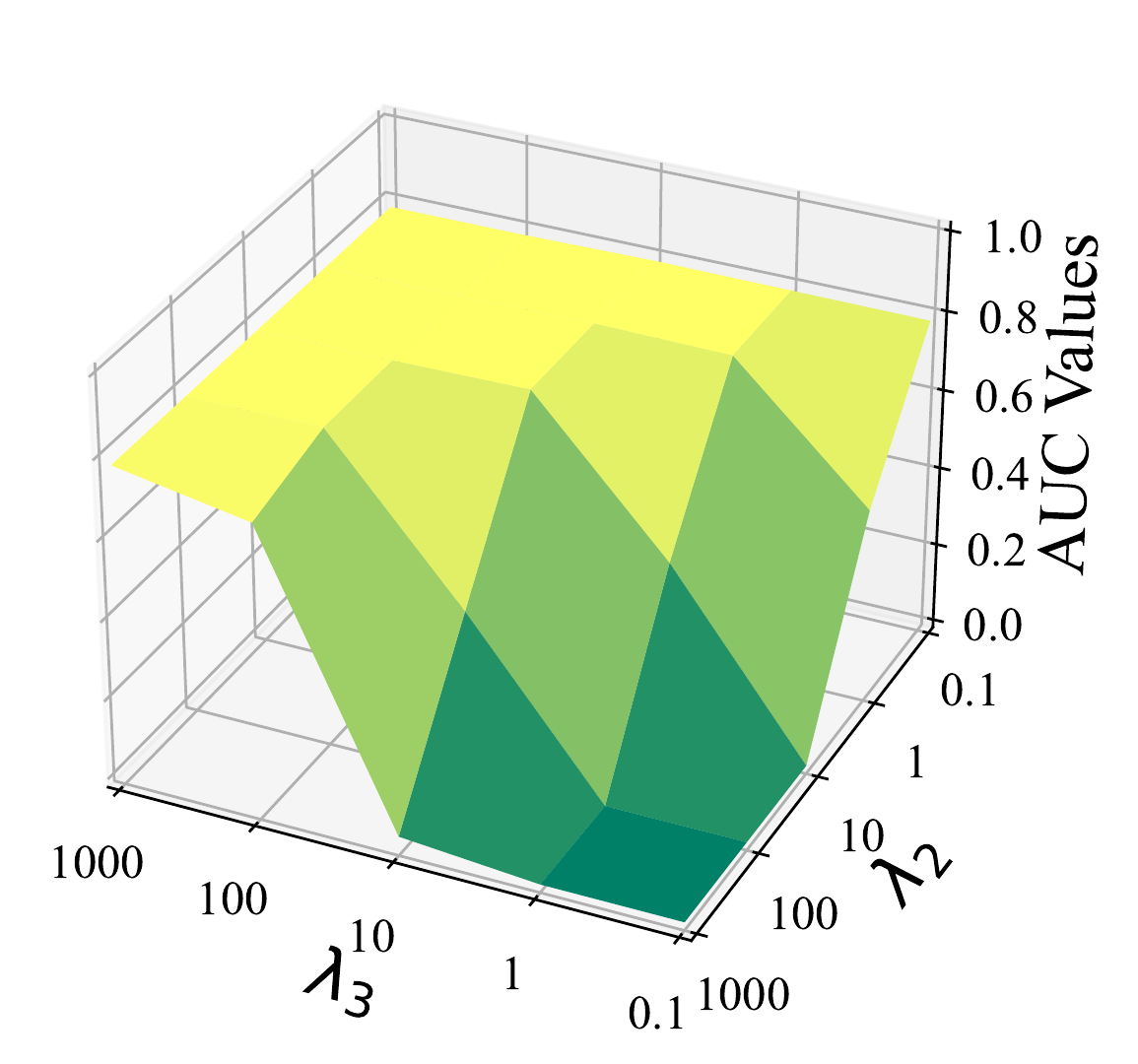}}
\subfigure[$\lambda_1=1000$]{
\includegraphics[width=0.18\linewidth]{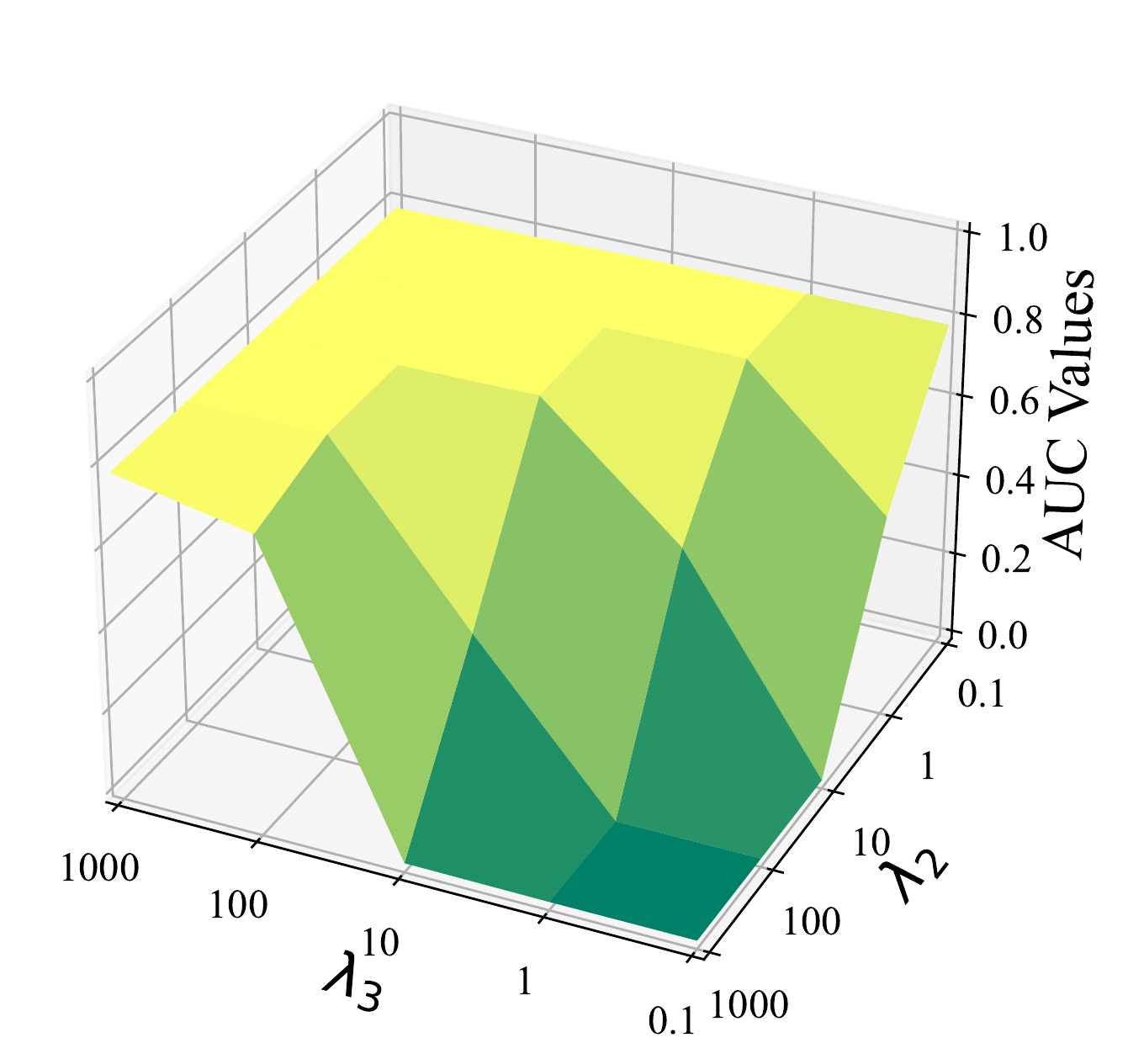}}
\caption{The AUC values of the SMSL model with respect to $\lambda_2$ and $\lambda_3$ under different $\lambda_1$ in (a)$-$(e) Object$\_$1550$\_$1558 dataset, (f)$-$(j) D1F12H1$\_$D1F12H2, and (k)$-$(o) D1F12H1$\_$D2F22H2.}
\label{para23}
\end{figure*}

It is known from Eq. (\ref{eq8}) that three trade-off parameters are utilized to balance the constraints of coefficient matrix $\mathrm{C}$ and $\mathrm{D}^s$ for the optimization procedure, where $\lambda_1$ reflects the influence of the consistency term $||\mathrm{C}||_\star$, $\lambda_2$ relates to the quantity of the specific matrices, and $\lambda_3$ reflects the orthogonality between the specific matrices. Considering that $\lambda_2$ and $\lambda_3$ are two penalty scalar on $\mathrm{\bf D}^s$, we jointly analyze the influence of $\lambda_2$ and $\lambda_3$ by setting their values as $\{0.1, 1, 10, 100, 1000\}$. The AUC values of the proposed SMSL method are illustrated by 3-D colored surfaces, as shown in Fig. \ref{para23}, where $\lambda_1$ is set to specific values. It can be seen that $\lambda_2$ and $\lambda_3$ are negatively correlated: when the value of $\lambda_2$ is smaller than $10$ and $\lambda_3$ is larger than $10$, the proposed model has stable and better performances. Our intuitive understanding is that paying more attention to the difference between the specific matrices and setting smaller weights on the sum of the specific matrices is helpful to detect anomalous changes. We can also observe in three datasets that the relationships between these two parameters are not much influenced by $\lambda_1$, which indicates that the parametric settings of the two specificity terms are independent of the consistency term. The promising result can be obtained when $\lambda_2=10$ and $\lambda_3=10$.

\begin{figure}
\centering
\includegraphics[width=0.78\linewidth]{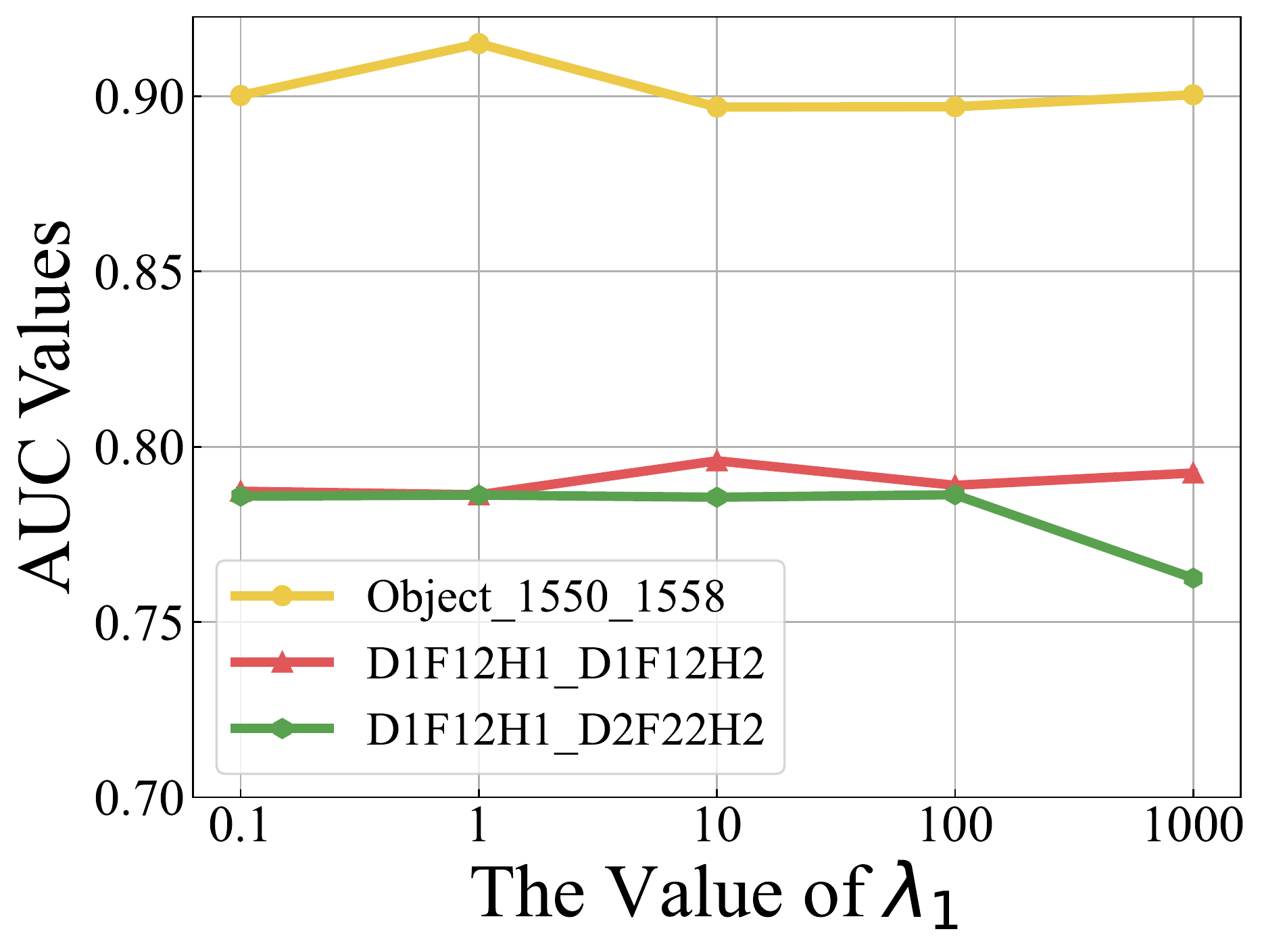}
\caption{The AUC values of the SMSL model with respect to $\lambda_1$ in three datasets, where $\lambda_2$ and $\lambda_3$ are fixed to 10.}
\label{para1}
\end{figure}

Then, we fix the values of $\lambda_2$ and $\lambda_3$ and observe the influence of $\lambda_1$ in three datasets. The results are shown in Fig. \ref{para1}. It can be found that as the value of $\lambda_1$ increasing, the performance of the SMSL model varies slightly. When the value of $\lambda_1$ equals 1, the SMSL model performs the best in the Object$\_$1550$\_$1558 dataset. For the Viareggio dataset, we can observe that the influence of $\lambda_1$ on the performance of the proposed model is not consistent, which may be caused by the different illumination conditions between two image pairs. For the ``D1F12H1$\_$D1F12H2'' image pair, we have the best result when $\lambda_1=10$. And the detection accuracy reaches the best when $\lambda_1=1$ in the ``D1F12H1$\_$D1F22H2'' image pair. Therefore, the recommended the range of $\lambda_1$ is $[1, 10]$ by experiencing.

\subsubsection{Convergence Study}
\begin{figure}[htp]
\centering
\includegraphics[width=0.75\linewidth]{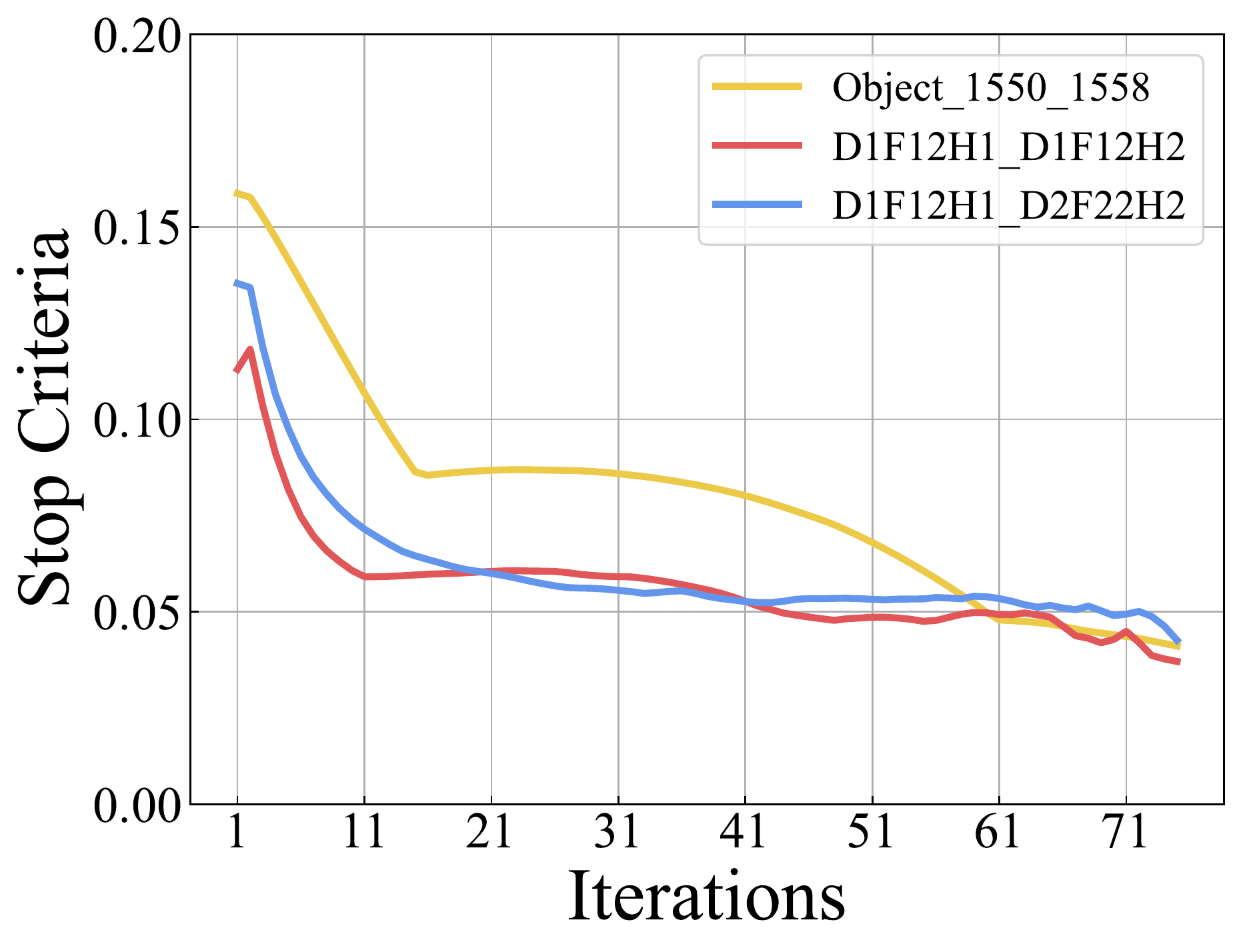}
\caption{The convergence curves about the overall stopping criteria of reconstruction errors versus the iterations on three datasets.}
\label{convergence}
\end{figure}

Finally, the convergence property of the proposed SMSL is explored on three datasets. As shown in Fig. \ref{convergence}, the stopping criteria related to the residuals at every iteration are given. It can be observed that the tendency of the SMSL is quickly converged at the first ten iterations, and the residuals keep decreasing in the subsequent iterations.

\section{Conclusion}
In this paper, which focuses on the hyperspectral anomalous change detection problem, we propose a sketched multi-view subspace learning (SMSL) model for large scale multi-temporal images. Unlike existing multi-view learning models that require redundant computational consumption and storage, the proposed SMSL model extracts the main information from the inputs and constructs a sketched dictionary for self-representation learning. In addition, the common information and the differences are simultaneously learned by defining a consistent part and specific parts for the coefficients. Further, the paper discussed a multiple sub-optimization process using the ALM algorithm in detail, and conducted sufficient experiments on large-scale hyperspectral datasets. The experimental results demonstrate the superiority of our approach compared to other classical methods of background suppression and anomalous changes extraction. Additionally, the parametric analysis with respect to the trade-off parameters and the convergence study of the model are also discussed.


\ifCLASSOPTIONcompsoc
  \section*{Acknowledgments}
\else
  \section*{Acknowledgment}
\fi

The authors would like to thank the handling editor and
the anonymous reviewers for their careful reading and helpful
remarks. The authors would also like to thank the Institute of Advanced Research in Artificial Intelligence (IARAI) for its support.

\ifCLASSOPTIONcaptionsoff
  \newpage
\fi

\bibliographystyle{IEEEtran}

\bibliography{ref}
%
\end{document}